\documentclass{article} 
\usepackage{iclr2026_conference,times}

\usepackage{amsmath,amsfonts,bm}

\def\eqref#1{equation~\ref{#1}}

\def\1{\bm{1}}

\DeclareMathAlphabet{\mathsfit}{\encodingdefault}{\sfdefault}{m}{sl}
\SetMathAlphabet{\mathsfit}{bold}{\encodingdefault}{\sfdefault}{bx}{n}

\usepackage{hyperref}
\usepackage{url}

\usepackage[utf8]{inputenc} 
\usepackage[T1]{fontenc}    
\usepackage{hyperref}       
\usepackage{url}            
\usepackage{booktabs}       
\usepackage{amsfonts}       
\usepackage{nicefrac}       
\usepackage{microtype}      
\usepackage{xcolor}         

\usepackage{makecell}
\usepackage{array}

\RequirePackage{fancyhdr}
\RequirePackage{algorithm}
\RequirePackage{algorithmic}
\RequirePackage{eso-pic} 
\RequirePackage{forloop}
\usepackage{amssymb}
\usepackage{mathtools}
\usepackage{amsthm}
\usepackage{cleveref}
\crefname{equation}{Eq.}{Eqs.}
\crefname{table}{Table}{Tables}
\crefname{figure}{Figure}{Figures}
\crefname{section}{Section}{Sections}
\crefname{algorithm}{Algorithm}{Algorithms}
\theoremstyle{plain}

\theoremstyle{definition}

\theoremstyle{remark}

\usepackage{colortbl}
\usepackage{adjustbox}
\usepackage{url}
\usepackage{multirow,multicol,xspace}
\usepackage{float}
\usepackage{graphics}
\usepackage{paralist}
\usepackage{wrapfig}
\usepackage[normalem]{ulem}
\usepackage[version=4]{mhchem}

\usepackage{pgfplots}
\pgfplotsset{width=8cm,compat=1.17} 
\usepackage{pifont}
\usepackage{subcaption,ragged2e}
\usepackage{caption}

\newcommand{\method}{{DemoDiff}\xspace}
\newcommand{\token}{{NPE}\xspace}

\usepackage{listings}
\title{Graph Diffusion Transformers are In-Context Molecular Designers}

\author{Gang Liu$^{1}$, \quad Jie Chen$^2$, \quad  Yihan Zhu$^{1}$ \quad  Michael Sun$^{3}$, \\ \bf Tengfei Luo$^{1}$, \quad  Nitesh V. Chawla$^{1}$, \quad Meng Jiang$^1$ \\
$^1$University of Notre Dame \quad $^2$ MIT-IBM Watson AI Lab, IBM Research \quad  $^3$MIT CSAIL \\
\texttt{\{gliu7, mjiang2\}@nd.edu}
}

\iclrfinalcopy
\begin{document}

\maketitle

\begin{abstract}
In-context learning allows large models to adapt to new tasks from a few demonstrations, but it has shown limited success in molecular design. Existing databases such as ChEMBL contain molecular properties spanning millions of biological assays, yet labeled data for each property remain scarce. To address this limitation, we introduce demonstration-conditioned diffusion models (\method), which define task contexts using a small set of molecule–score examples instead of text descriptions. These demonstrations guide a denoising Transformer to generate molecules aligned with target properties. 
For scalable pretraining, we develop a new molecular tokenizer with Node Pair Encoding that represents molecules at the motif level, requiring 5.5$\times$ fewer nodes. 
We curate a dataset containing millions of context tasks from multiple sources covering both drugs and materials, and pretrain a 0.7-billion-parameter model on it.
Across 33 design tasks in six categories, \method matches or surpasses language models 100–1000$\times$ larger and achieves an average rank of 3.63 compared to 5.25–10.20 for domain-specific approaches. These results position \method as a molecular foundation model for in-context molecular design.
\end{abstract}

\section{Introduction}
\label{sec:introduction}
In-context learning (ICL) is the emergent capability of large models to infer task-specific concepts from a few demonstrations~\citep{xie2021explanation}.  
ICL has been studied in large language models (LLMs), but was found less effective for molecular design than specialized methods~\citep{liu2024multimodal}. These specialized models often depend on extensive Oracle calls~\citep{gao2022sample} or large labeled datasets beyond what context examples provide. Molecular tasks, however, involve millions of types, many with only a few labeled examples~\citep{zdrazil2024chembl}. Such examples are enough to form task contexts but insufficient to train a new model.
This trade-off motivates our \textit{in-context molecular design}, which combines the flexibility of ICL with the efficiency of molecular domain knowledge. 

\begin{figure*}[ht]
    \centering
    \includegraphics[width=0.95\textwidth]{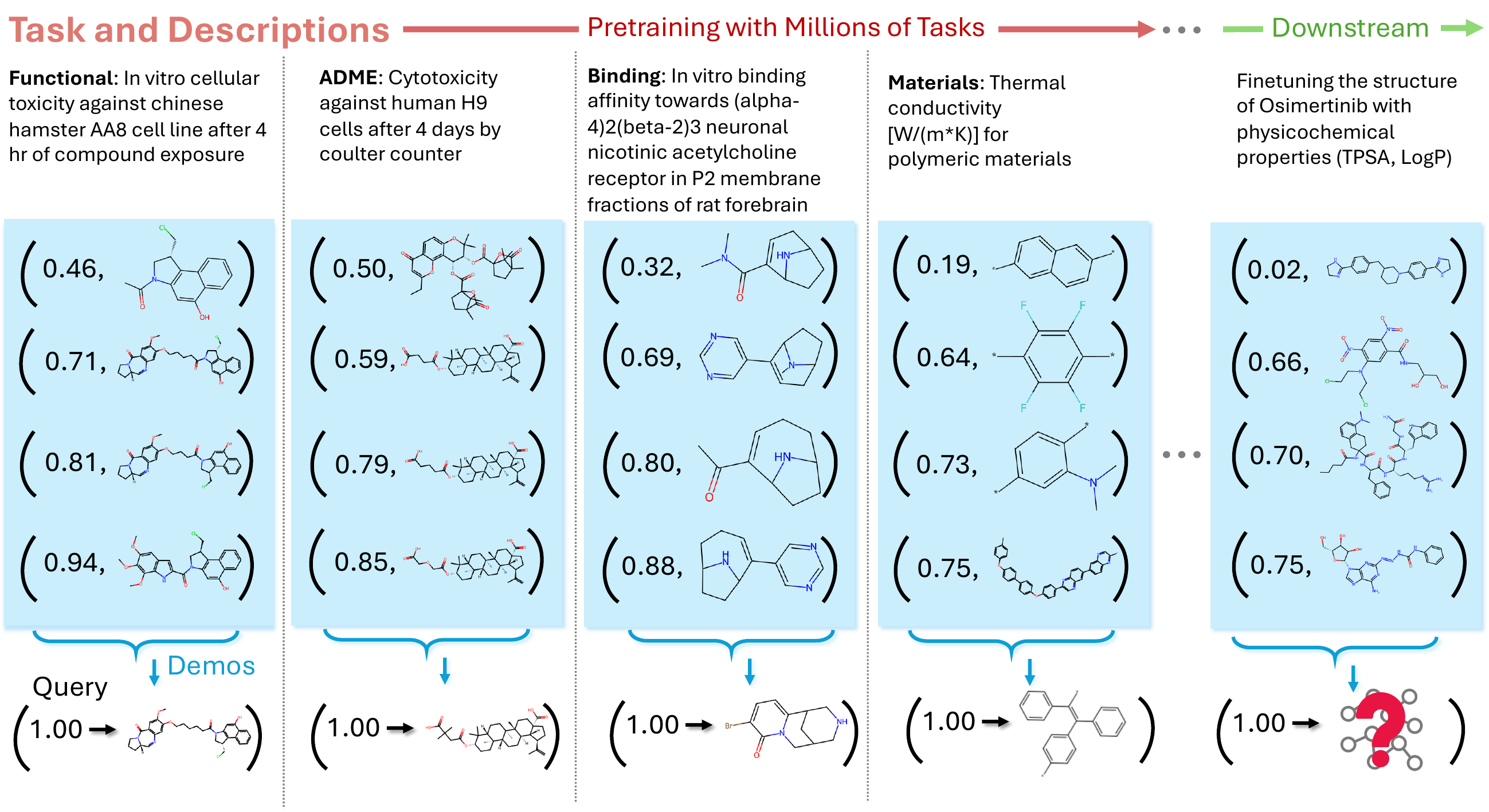}
    \vspace{-0.2in}
    \caption{In-context molecular design with \method. Each demo is defined as a score–molecule pair, and a set of them forms the task context as conditions. After pretraining on large and diverse tasks, \method serves as a foundation model for designing molecules in new task contexts. Scores represent relative distances to the target and are converted from raw labels, as shown in~\cref{subsec:pretrain}.}
    \label{fig:icl-idea}
    \vspace{-0.25in}
\end{figure*}

Molecular structures and properties are discrete graphs and numbers with varying scales and units. Directly adapting the autoregressive framework from LLMs is infeasible~\citep{brown2020language} for in-context molecular designs, as the input and output of language data are text in sequential order.
Diffusion models show promise for molecular structures~\citep{vignac2022digress}, and Graph Diffusion Transformers (Graph DiTs) are effective for modeling their joint distribution with properties~\citep{liu2024graph}. However, Graph DiTs have been studied with at most five properties represented in a single vector. In practice, molecular properties span millions of assays in biology, including functions, binding, ADME, and toxicity, as well as material properties such as gas permeability, thermal conductivity, and glass transition temperature (\cref{fig:icl-idea}). Representing all properties in one-hot vector with millions of dimensions is inefficient, produces sparse pretraining data since many assays have fewer than ten labels, and limits generalization to unseen properties in downstream tasks.

Instead of a property vector with a large embedding table, we use demonstrations to define the task context for molecular design. As shown in~\cref{fig:icl-idea}, the demonstrations consist of a set of molecules with scores in $[0,1]$ and molecular design is framed as a query for the target score of 1.
Molecules in the context do not follow a strict order, and their scores serve as \textit{relative} positions to the target, functioning as a replacement for position IDs in Transformers.
\citep{xie2021explanation} described ICL in LLMs as implicit Bayesian inference over latent \textit{concepts} expressed by examples in the prompt. Similarly, each task in~\cref{fig:icl-idea} shares the concepts defined by the joint distribution of molecules and their scores.
The denoising Transformer in the Graph DiT attends to the context, implicitly extracts concepts, and uses them to guide the reverse process to refine the structure. An example generation trajectory is shown in~\cref{fig:study-trajectory}.
A simple way to represent the task concept is to use positive demonstrations, such as molecules close to the target or active in the assay. However, positive examples alone may be insufficient, as they can overlap across tasks due to factors such as task relatedness (e.g., activity in non-small-cell lung cancer but across different cell lines) or due to sampling bias when the set of positives is extremely sparse (e.g., only one positive example shared by two tasks). To address this, we form the task context using not only positive but also medium and negative examples, providing a more complete representation of the task concepts.

With these task contexts, we propose \textbf{demo}nstration-conditioned \textbf{diff}usion models (\method) and pretrain a 0.7B model with a Graph DiT as the backbone, using over 140 H100 GPU days. 
To support efficient pretraining, we introduce a molecular tokenizer trained with Node Pair Encoding (NPE) for motif-level representation. On average, it reduces the number of nodes by 5.5× compared to atom-level representations (\cref{fig:num-node-full,fig:compress-ratio}).
The tokenizer iteratively merges neighboring nodes and selects frequent motifs to construct vocabularies. Motifs are connected by directed edges that preserve bond types and attachment rules, ensuring lossless reconstruction. Graph DiTs naturally use this motif-level representation and attend to motif semantics for denoising. 
For pretraining, we construct a dataset of over 1.6 million tasks from 155K unique properties and one million molecules. It combines ChEMBL for drugs~\citep{zdrazil2024chembl} and multiple polymer data sources for materials~\citep{otsuka2011polyinfo,thornton2012polymer,kuenneth2021polymer}.
For ICL, we propose a consistency score as a confidence measure of whether a generation aligns more closely with higher-scoring molecules in demonstrations, effectively filtering out false positives in generation (\cref{subsec:case-study}).

We evaluate \method-0.7B on 33 design tasks across six categories. It matches or surpasses LLMs 100–1000$\times$ larger in generating diverse, high-scoring molecules. Compared to ten specialized models (average ranks 5.25–10.20), \method ranks 3.6, demonstrating its strength as a molecular foundation model. The new molecular tokenizer further improves representation efficiency.

\section{Preliminaries}
\vspace{-0.1in}
\subsection{In-Context Learning with Demonstrations}

In ICL, the context is a set of demonstrations $\mathcal{C} = \{e_i\}_{i=1}^L$, where each $e_i$ is an input–label pair. Following~\citep{xie2021explanation}, we assume $\mathcal{C}$ reflects a latent concept $\theta \in \Theta$ from a family of concepts $\Theta$. For example, for a paragraph about Albert Einstein, the latent concept may be biography. Given a query $Q$, a foundation model with ICL generates an output $X$ by marginalizing over $\theta$:
\begin{equation}\label{eq-prob-icl} 
    p(X \mid \mathcal{C}, Q) = \int_{\theta} p(X \mid \theta, \mathcal{C}, Q)\, p(\theta \mid \mathcal{C}, Q)\, d\theta.
\end{equation}
Here $(\mathcal{C}, Q)$ form the prompt. If $p(\theta | \mathcal{C}, Q)$ concentrates on the prompt concept with more demonstrations, then the model identifies and applies that concept through marginalization. ICL can thus be implicit Bayesian inference.
All context, query, and outcomes are texts in language modeling. In inverse molecular design, we have molecule-score pairs as demonstrations. They capture the latent task concept, with semantics like the task descriptions in~\cref{fig:icl-idea}. $Q$ is the target score, and $X$ is the molecule to be designed. We focus on a new molecular foundation model for ICL.

\subsection{Molecular Design with Graph Diffusion Transformers}\label{subsec:prepare-graphdit}

Molecules are discrete graphs $X = (A, B)$, where $A$ denotes the set of atoms and $B$ the set of bonds. These structures are commonly modeled using discrete diffusion processes~\citep{vignac2022digress,liu2024graph}. Graph DiTs concatenate atom and bond features to $X$ into the input format of standard Transformers.
Given $X$, for each atom $a_i \in A$ with $d_i$ neighbors, Graph DiTs define a token as $x = \{a_i, \{b_{ij}\}_{j=1}^{d_i}\}$, where $b_{ij} \in B$ encodes the bond type (single, double, triple, or none). Each token is represented by a feature vector $\mathbf{x} \in \mathbb{R}^{F}$, formed by concatenating the one-hot encoding of the atom type and the connection types to all other atoms (either a bond type or a null type indicating no connection).
Discrete diffusion has a transition matrix $\mathbf{Q}$, initialized based on the frequency of atoms and bonds in the training set. At step $t$, $[\mathbf{Q}^t]_{ij} = q(\mathbf{x}^t_j \mid \mathbf{x}^{t-1}_i)$ for $i,j \in [1, F]$.

The forward diffusion with $\mathbf{Q}$ is: $q(\mathbf{x}^{t} \mid \mathbf{x}^{t-1}) = \operatorname{Cat}(\mathbf{x}_t; \mathbf{p}=\mathbf{x}^{t-1}\mathbf{Q}^{t})$, where $\operatorname{Cat}(\mathbf{x}; \mathbf{p})$ denotes two separate categorical sampling for atoms and bonds with probabilities from $\mathbf{p}$.
Starting from the original data $\mathbf{x}=\mathbf{x}^0$, we have $q(\mathbf{x}^t \mid \mathbf{x}^0) = \operatorname{Cat}\left(\mathbf{x}^t; \mathbf{p}=\mathbf{x}^0 \bar{\mathbf{Q}}^t\right)$, where $\bar{\mathbf{Q}}^t = \prod_{i\leq t} \mathbf{Q}^{i}$. The forward diffusion gradually corrupts data points. When the total timestep $T$ is large enough, $q(\mathbf{x}^T)$ converges to a stationary distribution.
The reverse process samples from $q(\mathbf{x}^T)$ and gradually removes noise. The posterior distribution $q(\mathbf{x}^{t-1} \mid \mathbf{x}^{t})$ is calculated as $q(\mathbf{x}^{t-1}|\mathbf{x}^t, \mathbf{x}^0) \propto \mathbf{x}^t (\mathbf{Q}^t)^\top \odot \mathbf{x}^0 \bar{\mathbf{Q}}^{t-1}$. 
Given multiple properties $\{c_i\}_{i=1}^n$, the denoising model approximates $p_{\phi}(\mathbf{x}^{t-1} \mid \mathbf{x}^t, \mathbf{x}^0, \{c_i\}_{i=1}^n)$ under property conditions. This model is trained by minimizing the negative log-likelihood for $\mathbf{x}^0$:
\begin{equation}\label{eq-discrete-diffusion-loss}
\mathcal{L}_\text{DM} = \mathbb{E}_{q(\mathbf{x}^0)} \mathbb{E}_{q(\mathbf{x}^t \mid \mathbf{x}^0)} \left[-\log p_{\phi} \left( \mathbf{x}^0 \mid \mathbf{x}^t, c_1, c_2, \dots, c_n \right) \right],
\end{equation}
In Graph DiTs~\citep{liu2024graph}, the constraints $\{c_i\}_{i=1}^n$ are numerical or categorical property values. In this work, we explore them as demonstrations for in-context learning.

\section{Learning Diffusion Model with Demonstrations}
\begin{figure*}[t]
    \centering
    \includegraphics[width=0.95\textwidth]{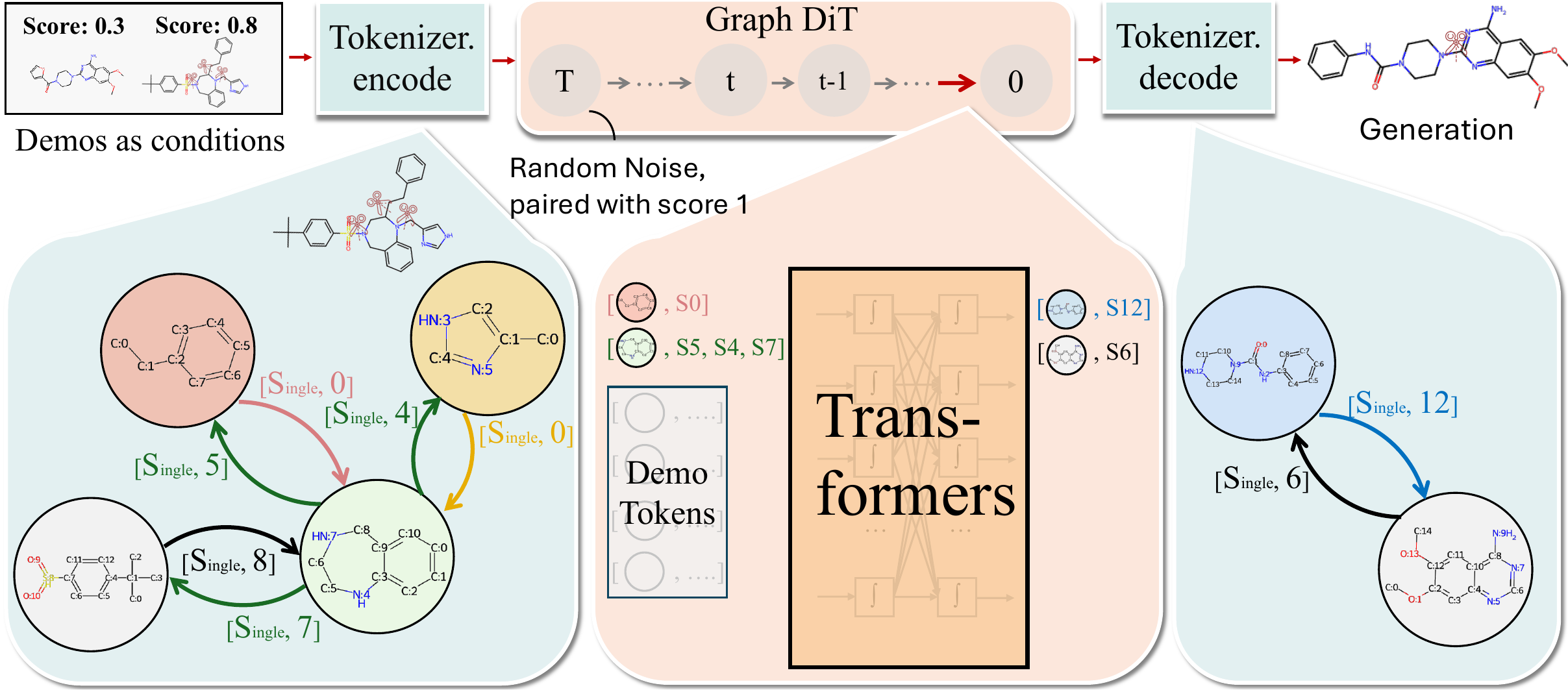}
    \caption{Demonstration-conditioned diffusion generation. In the reverse process, \method starts from random noise and denoises molecules conditioned on a set of molecule–score demonstration pairs at the motif level, with a tokenizer bridging motif and atom representations.
    }
    \label{fig:overview-method}
\end{figure*}

\cref{fig:overview-method} shows the generation process of \method, combining motif-based representation (\cref{subsec:tokenizer}) with graph diffusion transformers for in-context molecular generation (\cref{subsec:icl-gdit}).

\subsection{Molecular Graph Tokenization with Node Pair Encoding}\label{subsec:tokenizer}

More demonstrations help capture the latent task concept and are empirically useful~\citep{bertsch2024context}. However, prior work~\citep{liu2024graph} uses atom-level molecular representations, similar to modeling text at the character level, which fundamentally limits the number of examples in context.
For efficient representation, we merge frequent sub-molecular patterns as \textit{motif} $m = (\tilde{A}, \tilde{B}) \subseteq X$, where $\tilde{A} \subseteq A$ and $\tilde{B} \subseteq B$ define a connected substructure.
A molecule becomes a collection of disjoint motifs $M = \{m_i\}_{i=1}^{n}$ such that: (1) $\tilde{A}_i \cap \tilde{A}_j = \emptyset$ for all $i \neq j$; (2) $\bigcup_{i=1}^{n} \tilde{A}_i = A$.
Then, each edge $e_{ij} \in E$ is directed from a source motif $m_i$ to a target motif $m_j$, with two associated attributes: (1) bond type, and (2) attachment specification, indicating the atom within $m_i$ from which the bond originates. This abstraction induces a tokenizer with two functions. They are $\operatorname{tokenizer.encode} : X = (A, B) \longmapsto \hat{X} = (M, E)$ that compresses the atom-level graph into motif-level form and $\operatorname{tokenizer.decode} : \hat{X} = (M, E) \longmapsto X = (A, B) $ for reconstruction. The tokenizer uses two vocabularies: $\mathcal{M}$ for motif types and $\mathcal{E}$ for edge types. It starts with the 118 atom types from the periodic table and one ``*'' for the polymerization point, which form the initial motifs in $\mathcal{M}$. This guarantees that, in the worst case, a new molecule can still be represented at the atom level.
In each iteration, the tokenizer merges the most frequent neighbors until no further merge is found in $\mathcal{M}$, then proceeds with the corresponding connections between motifs.

To construct $\mathcal{M}$, existing methods, such as BRICS~\citep{degen2008art} or molecular grammars \citep{sun2025foundation, sun2024representing}, rely on domain-specific heuristics based on chemical reactions or expert knowledge. The resulting vocabularies are independent of the pretraining data, often missing frequent motifs.
To address this limitation, we propose \textit{Node Pair Encoding} (NPE), a frequency-based algorithm for molecular graphs inspired by BPE, as outlined in~\cref{alg:npe-merge}. We initialize $\mathcal{M}$ with elements from the periodic table and polymerization points ``*''. NPE iteratively performs three steps:  
(1) \textbf{Neighborhood merge}: For each molecule $X \in \mathcal{D}$ in the dataset $\mathcal{D}$ and current $\mathcal{M}$, we identify candidate motifs by merging adjacent substructures that appear in $\mathcal{M}$;
(2) \textbf{Frequency selection}: The most frequent candidate motif is selected and added to $\mathcal{M}$;  
(3) \textbf{Graph update}: Each $X \in \mathcal{D}$ is updated by replacing instances of merged motif pairs with the new motif.

\textbf{Constrained NPE}: 
The standard NPE may produce multiple directed edges from a motif $m_i$ to another $m_j$ when decomposing ring structures (e.g., aromatic rings). It leads to ambiguity during decoding since $e_{ij}$ does not uniquely determine the attachment specification within $m_j$. To address this, we introduce constraints such as rings into NPE at two stages.
During initialization, we traverse each molecule to identify its set of maximal connected rings, denoted $\mathcal{R}$, compute their frequencies, and include the top-$K_{\text{ring}}$ most frequent rings in the initial vocabulary $\mathcal{M}$. During motif merging, any $m \in \mathcal{M}$ is merged with a ring $r \in \mathcal{R} \setminus \mathcal{M}$ as a complete unit, rather than merging individual atoms within $r$. This strategy integrates frequent rings into the vocabulary while preserving atom-level representations for rare rings, avoiding reconstruction ambiguity.

Using \token, a molecule is represented as $n$ tokens $\{x_i\}_{i=1}^{n}$, where $x_i = \{m_i, \{e_{ij}\}_{j=1}^{d_i}\}$, with $m_i$ denoting a motif and $e_{ij}$ the associated edges. We set the motif vocabulary size to $K=3000$ ($K_{\text{ring}}=300$), with details and analysis provided in~\cref{addsec:tokenizer-prepare}.
As shown in~\cref{fig:overview-method}, an example input with 38 atoms can be compactly expressed using four motifs. An empirical comparison of atom- and motif-level representations over 1 million pretraining molecules is given in~\cref{fig:atom-motif}, with an average compression ratio of $5.446 \pm 2.569$, reducing the median count from 30 atoms to 5 motifs.

\subsection{In-Context Learning with Graph Diffusion Transformers}\label{subsec:icl-gdit}

We construct the dataset $\mathcal{D} = \{(\mathcal{C}_i, Q_i, X_i)\}_{i=1}^{N_{\text{pretrain}}}$ for pretraining, where each task consists of a context of molecule–score pairs $\mathcal{C}_i$, a query score $Q_i$, and a target molecule $X_i$.
To pretrain \method for in-context inverse molecular design, we replace the property conditions in~\cref{eq-discrete-diffusion-loss} with $\mathcal{C}$ and $Q$:
\begin{equation}\label{eq-icl-pretrain}
\mathcal{L}_\text{pretrain} = \mathbb{E}_{q(\mathbf{x}^0)} \mathbb{E}_{q(\mathbf{x}^t \mid \mathbf{x}^0)} \left[-\log p_{\theta} \left( \mathbf{x}^0 \mid \mathbf{x}^t, \mathcal{C}, Q \right) \right].
\end{equation}
With large and diverse pretraining data and scalable Transformers~\citep{peebles2023scalable}, \method learns to infer the latent task concept to generate the target molecule and can serve as a foundation model for ICL.
In a task, it performs implicit Bayesian inference over diffusion trajectories.

\paragraph{ICL with Context Consistency (\cref{addsec:consistency-score}):} 
Given a query $Y$, demonstrations $\mathcal{C}_i$ are divided into positive, medium, and negative groups. For a generated molecule $X$, we compare its fingerprint-based similarity with these groups to assess whether it follows the relation $\mathrm{pos} > \mathrm{med} > \mathrm{neg}$. This yields a consistency score that measures how well the generated molecule aligns with the relative relations in the context. In experiments (\cref{sec:perf-discussion}), we use this score to select high consistent generations before conducting the final evaluation with Oracles.

\subsection{Model Design and Pretraining}\label{subsec:pretrain}

\begin{figure}[ht]
    \centering
    \begin{subfigure}[b]{0.48\textwidth}
        \centering
        \includegraphics[width=\textwidth]{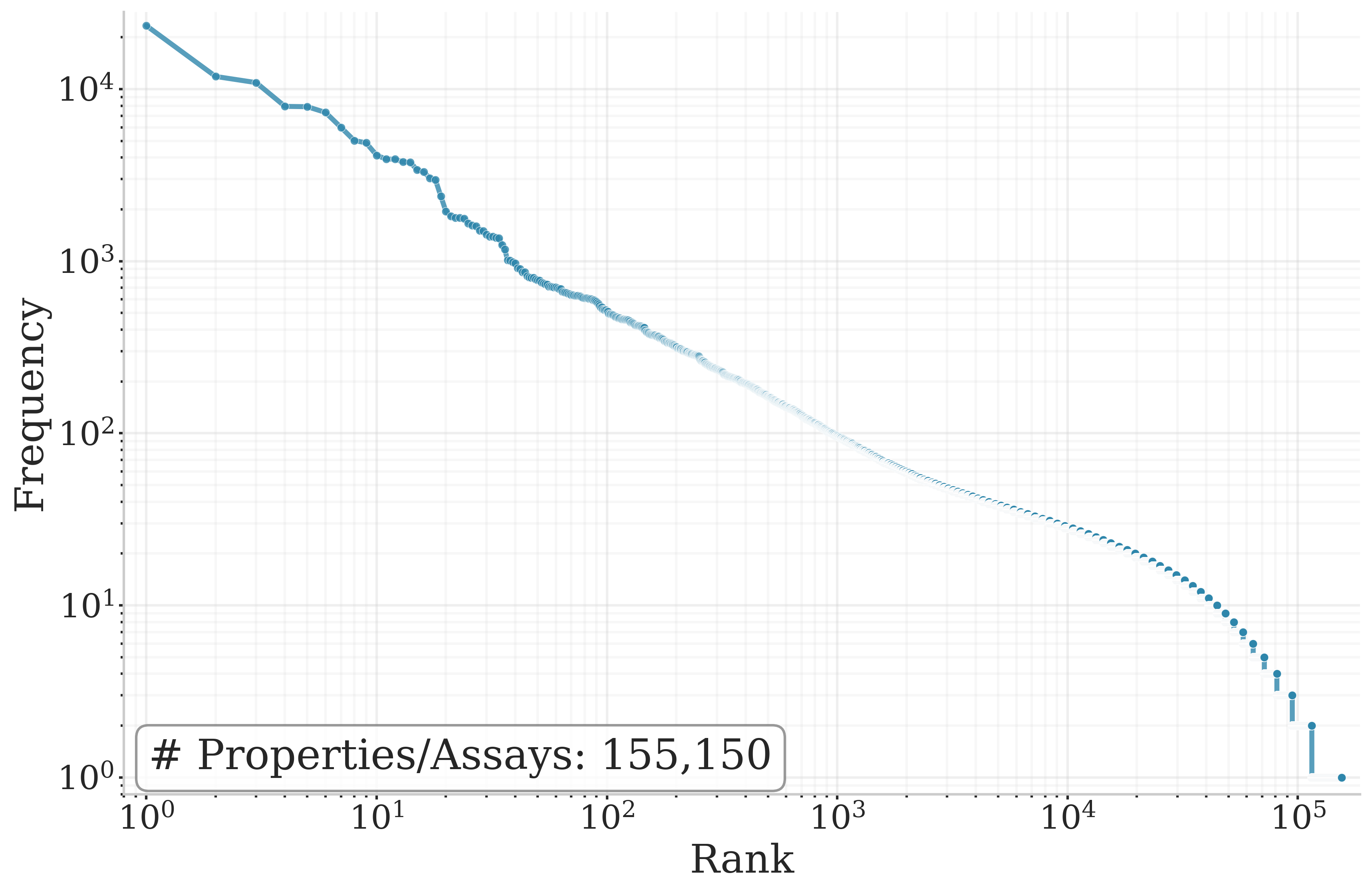}
        \caption{Property rank vs. frequency in $\mathcal{D}$. 
        }
        \label{fig:task-freq}
    \end{subfigure}
    \hfill
    \begin{subfigure}[b]{0.48\textwidth}
        \centering
        \includegraphics[width=\textwidth]{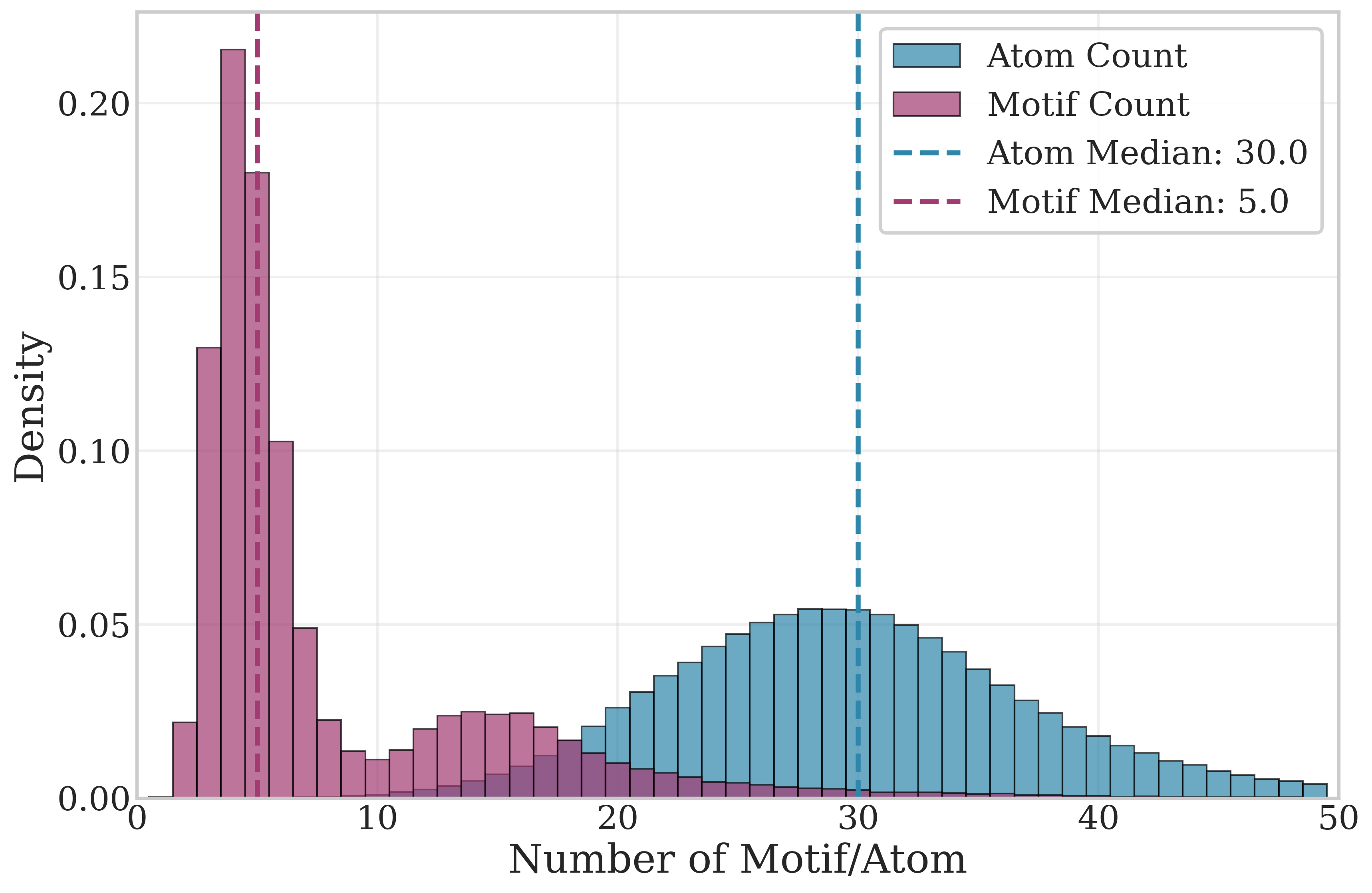}
        \caption{Node counts ($n \leq 50$; full in~\cref{fig:num-node-full}).}
        \label{fig:atom-motif}
    \end{subfigure}
    \caption{Pretraining data statistics for property rank-frequency and node count density.}
    \label{fig:pretrain-stats}
\end{figure}

\textbf{Model Designs:} 
\cref{fig:overview-method} illustrates the model architecture. For the $i$-th task in the pretraining set $(\mathcal{C}_i, X_i, Y_i) \in \mathcal{D}$, the property score $Y_i$ is scaled within $[0, 1]$, providing positional signals for both demonstration molecules and the target. These scalar values are encoded using Rotary Position Embedding (RoPE)~\citep{su2024roformer}.
\method uses a tokenizer to process the atom-level representation and defines a maximum context length for the number of motif tokens. The context includes the target along with as many demonstration tokens as fit within the target length. Since molecules in the context are structurally disjoint, edge connectivity implicitly delineates context boundaries, removing the need for explicit delimiter tokens. Details are in~\cref{addsec:model-details}.

\textbf{Pretraining:} 
To construct the pretraining dataset as illustrated in~\cref{subsec:icl-gdit}, we use the ChEMBL database~\citep{zdrazil2024chembl}, the largest collection of biological assays, containing over 2.5 million molecules and 1.7 million assay records.
To increase chemical diversity for materials discovery, we augment ChEMBL with polymer datasets from multiple sources~\citep{liu2024multimodal,kuenneth2021polymer}, including properties such as thermal conductivity, free volume fraction, and glass transition temperature.
For biological assays, we generate tasks by selecting a molecule-assay pair as the target and treating other molecule-assay pairs as context. The target is assigned a score of 1, and context scores are computed by normalizing differences in $p\mathrm{ChEMBL}$ values (negative log of bioactivity measures such as IC50 and potency) to the interval $[0, 1]$. We restrict targets to bioactive molecules with $p\mathrm{ChEMBL}$ $> 6$. For polymers, we apply the same strategy: each polymer is used as a target, and its property value is normalized against those of other polymers to form context-target pairs. We partition context examples into three groups by normalized scores: positive $[0.75,1]$, medium $(0.5,0.75]$, and negative $[0,0.5]$, with up to 15 demonstrations from each.
The final dataset comprises around 1 million molecules with 155K unique assays or properties, yielding 1.6 million tasks. As shown in~\cref{fig:task-freq}, the frequency distribution of assays and properties follows Zipf's law, $P(Y_{\mathrm{rank}}) \propto \mathrm{rank}^{-1.13}$, consistent with patterns in language corpora. These 1 million molecules are used to initialize the motif vocabulary via \token, which provides a more compact representation by reducing node counts (\cref{fig:atom-motif}). We further extract edge connections to construct an edge vocabulary capturing motif-to-motif connectivity. 
Finally, we pretrain a \method model with 0.7B parameter on~\cref{eq-icl-pretrain}, using 146 H100 GPU days. Details are provided in~\cref{addsec:pretraining-details}.

\section{Experiment}\label{sec:experiment}
\textbf{Setups:}
We curate 33 downstream tasks (see~\cref{tab:overall_top10_harmonic,addsec:setups}) across six categories to evaluate \method against 13 baselines. These tasks are primarily curated by domain experts and are distinct from pretraining. We include eight molecular optimization methods and two conditional generation models (LSTMs and Graph DiT~\citep{liu2024graph}), and LLMs (DeepSeek-V3, GPT-4o, and Qwen-Max). 
We select the top four molecular optimization algorithms from the PMO benchmark~\citep{gao2022sample} (out of 25 evaluated methods), under two settings: 100 oracle calls and 10,000 predictor calls. 
For evaluation, we generate 10 valid, unique, and novel molecules per task and score them with Oracles. We report the harmonic mean over two dimensions: (a) averaged oracle scores and (b) the diversity score~\cref{eq:int-div}.
Each task has up to 450 molecule–score pairs, evenly divided into positive $[0.75,1]$, medium $(0.5,0.75]$, and negative $[0,0.5]$ groups. 
Each task has an Oracle function for evaluation, with limited budgets for Oracle calls. We use all molecules to train the task-specific predictor for predictor calls or to train conditional generation models directly. For different ICL methods, demonstrations are randomly sampled with a similar budget for context.

\subsection{Performance on Diverse Molecular Designs Tasks}~\label{sec:perf-discussion}
\begin{table}[t]
\centering
\caption{Harmonic mean of oracle and diversity scores. We group 33 tasks into six categories and report the mean ± std within each category. The best results in each column are \textbf{bolded}. Task-specific results and additional metrics are provided in~\cref{addsec:disc-result}.
}
\label{tab:overall_top10_harmonic}
\resizebox{\textwidth}{!}{%
\begin{tabular}{lcccccccc}
\toprule
Task Category & Drug & Drug & Structure & Drug & Target & Material & Avg & Total \\
 & Rediscovery & MPO & Constrained & Design & Based & Design & Rank & Sum \\
\# Tasks & 7 & 7 & 5 & 4 & 5 & 5 & 33 & 33 \\
\midrule
\multicolumn{9}{c}{Molecular Optimization Methods with 100 Oracle Calls} \\
\midrule
GraphGA & 0.36±0.07 & 0.52±0.19 & 0.43±0.21 & 0.41±0.32 & 0.76±0.04 & 0.58±0.11 & 5.25 & 16.65 \\
REINVENT & 0.37±0.08 & 0.52±0.17 & 0.43±0.21 & 0.42±0.32 & 0.76±0.03 & 0.00±0.00 & 6.70 & 13.84 \\
GPBO & 0.37±0.07 & 0.51±0.18 & 0.42±0.22 & 0.39±0.33 & 0.76±0.03 & 0.60±0.21 & 5.63 & 16.65 \\
STONED & 0.36±0.07 & 0.52±0.19 & 0.43±0.21 & 0.41±0.32 & 0.76±0.04 & \scriptsize NO SELFIES & 6.42 & 13.75 \\
\midrule
\multicolumn{9}{c}{Molecular Optimization Methods with 10000 Predictor Calls} \\
\midrule
GraphGA & 0.37±0.09 & 0.50±0.18 & 0.45±0.29 & 0.49±0.26 & 0.64±0.06 & 0.55±0.14 & 7.34 & 16.30 \\
REINVENT & 0.30±0.13 & 0.23±0.23 & 0.25±0.24 & 0.38±0.26 & 0.17±0.13 & 0.51±0.16 & 10.20 & 9.82 \\
GPBO & 0.33±0.10 & 0.45±0.22 & 0.43±0.29 & 0.49±0.15 & 0.74±0.03 & 0.42±0.24 & 7.80 & 15.35 \\
STONED & 0.33±0.10 & 0.40±0.20 & 0.50±0.28 & 0.27±0.10 & 0.20±0.27 & \scriptsize NO SELFIES & 9.24 & 9.66 \\
\midrule
\multicolumn{9}{c}{Conditional Generation Models} \\
\midrule
LSTM & 0.39±0.25 & 0.16±0.07 & 0.55±0.32 & 0.33±0.35 & 0.72±0.04 & 0.16±0.11 & 9.31 & 12.30 \\
Graph-DiT & 0.43±0.21 & 0.50±0.18 & \textbf{0.58±0.34} & 0.48±0.37 & 0.71±0.04 & 0.55±0.17 & 6.91 & 17.64 \\
\midrule
\multicolumn{9}{c}{Learning from In-Context Demonstrations} \\
\midrule
DeepSeek-V3 & 0.45±0.18 & 0.51±0.20 & 0.49±0.24 & 0.65±0.18 & 0.64±0.06 & 0.39±0.24 & 6.76 & 16.90 \\
GPT-4o & \textbf{0.47±0.21} & 0.53±0.20 & 0.52±0.30 & 0.48±0.40 & 0.73±0.05 & 0.43±0.16 & 6.34 & 17.25 \\
Qwen-Max & 0.15±0.21 & 0.17±0.15 & 0.32±0.32 & 0.29±0.29 & 0.19±0.26 & 0.10±0.18 & 11.56 & 6.46 \\
DemoDiff (Ours) & 0.44±0.21 & \textbf{0.54±0.23} & 0.56±0.33 & \textbf{0.79±0.11} & \textbf{0.78±0.05} & \textbf{0.67±0.11} & \textbf{3.63} & \textbf{20.10} \\
\bottomrule
\end{tabular}
}
\end{table}

\textbf{ICL achieves competitive performance with minimal supervision.} 
\cref{tab:overall_top10_harmonic} compares the harmonic mean of the top-10 generated molecules based on both task scores and structural diversity, while \cref{tab:overall_top1} (appendix) reports the top-1 scoring molecule per task.
Under limited data and Oracle budgets, ICL methods perform comparably to, or better than, fully trained conditional generators and molecular optimization baselines. 
Excluding Qwen-Max, \method and other LLM-based ICL approaches consistently attain top-tier average ranks. These ICL methods rely on tens of demonstrations per task, significantly fewer than the training data or Oracle calls required by other models or algorithms.

\textbf{\method designs molecules with accurate scores and high diversity.} 
Across six task categories, it performs best on property-driven tasks, including drug design with bioactivity targets, protein binding affinity, and material design for polymer gas separation. It achieves the lowest average rank of 3.62, outperforming the best baseline, GraphGA (rank 5.25). ICL methods with LLMs produce high-scoring top designs (\cref{tab:overall_top1}) but often generate structurally similar molecules. These do not necessarily align better with the target score while reducing diversity. In contrast, \method designs molecules with scores closer to the query and better structural diversity.

\textbf{\method performs better on property-driven tasks than on structure-constrained ones.} 
It scores 0.67–0.79 on drug and material design, but around 0.44–0.56 for rediscovery and structure-constrained tasks, where Oracle scoring is tied to the presence of specific structures. While \method still ranks highly in structure-constrained tasks, its stronger results on property-driven tasks highlight its advantage in exploring chemical spaces with broader solution ranges.

\subsection{Ablation Studies and Performance Analysis}
\begin{table}[htbp]
\centering
\caption{Performance across model sizes using harmonic mean scores from Top-10 generations}
\label{tab:perf-vs-size}
\resizebox{0.9\textwidth}{!}{%
\begin{tabular}{lcccccc}
\toprule
\makecell{\method} & \makecell{Drug \\ Rediscovery} & \makecell{Drug \\ MPO} & \makecell{Structure \\ Constrained} & \makecell{Drug \\ Design} & \makecell{Target \\ Based} & \makecell{Material \\ Design} \\
\midrule
78M & 0.39 ± 0.17 & 0.46 ± 0.24 & 0.59 ± 0.06 & 0.57 ± 0.31 & 0.73 ± 0.03 & 0.62 ± 0.13 \\
311M & 0.40 ± 0.17 & 0.46 ± 0.23 & 0.63 ± 0.06 & 0.53 ± 0.27 & 0.75 ± 0.04 & 0.62 ± 0.14 \\
739M & 0.44 ± 0.21 & 0.54 ± 0.23 & 0.56 ± 0.33 & 0.79 ± 0.11 & 0.78 ± 0.05 & 0.67 ± 0.11 \\
\bottomrule
\end{tabular}
}
\end{table}

\textbf{Model Parameters:} 
We pretrain \method with varying sizes: small (78.7M), medium (311M), and large (739M) parameters. \cref{tab:perf-vs-size} reports performance using the top-10 harmonic means of task score and diversity. We present averages with deviations across six categories. \method achieves reasonable scores even at small scale. At the medium scale, performance improves in most tasks except drug design, while the benefits of parameter scaling become more evident at the large scale.

\begin{figure}[t]
    \centering
    \resizebox{0.8\textwidth}{!}{%
        \begin{subfigure}[b]{0.4\textwidth}
            \centering
            \includegraphics[width=\textwidth]{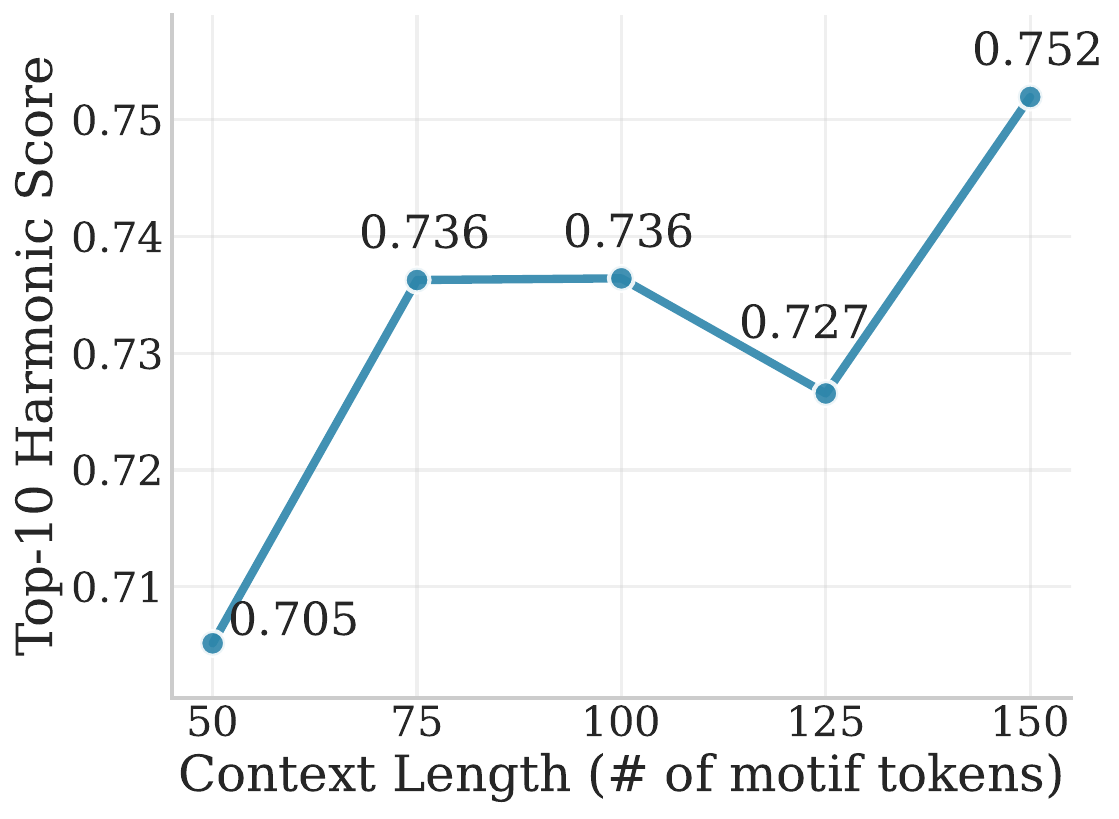}
            \caption{Harmonic score vs. context length}
            \label{fig:albuterol-contextlen}
        \end{subfigure}
        \hfill
        \begin{subfigure}[b]{0.4\textwidth}
            \centering
            \includegraphics[width=\textwidth]{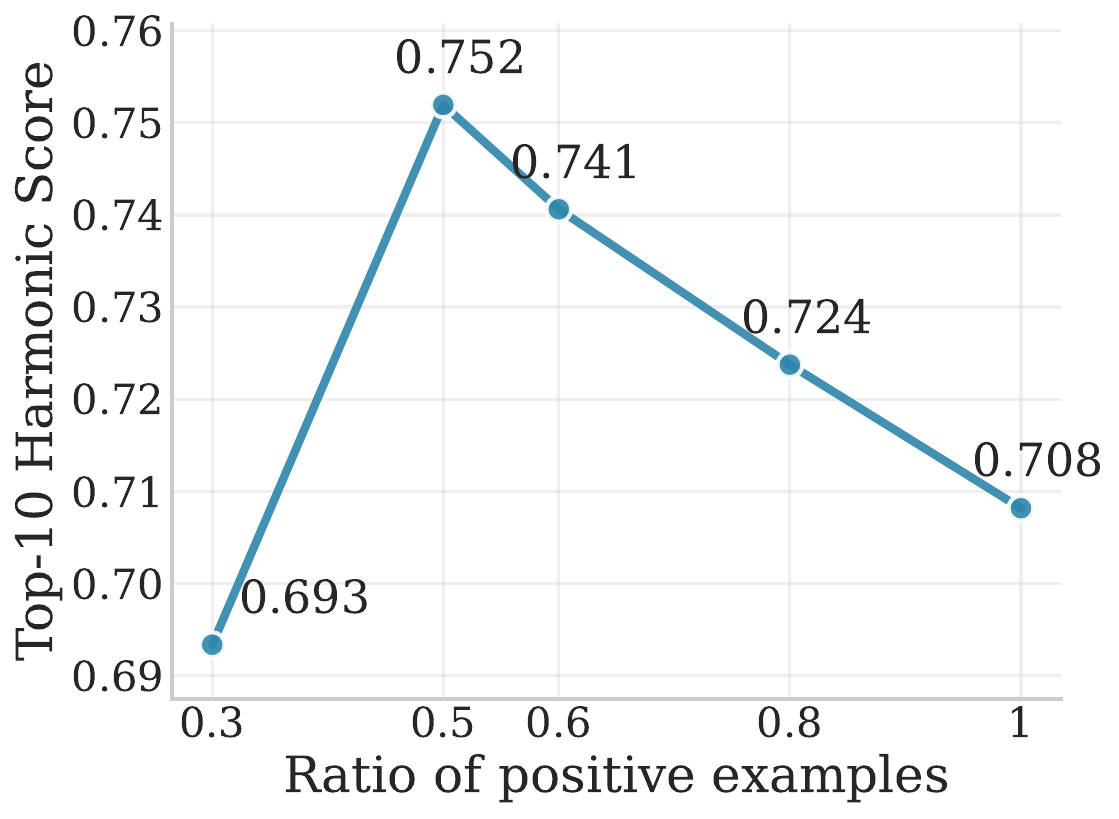}
            \caption{Harmonic score vs. positive samples}
            \label{fig:albuterol-pos}
        \end{subfigure}
    }
    \caption{Ablation studies on Albuterol drug rediscovery.}
    \label{fig:ablate-albuterol}
\end{figure}

\textbf{ICL with Demonstrations:} 
\cref{fig:ablate-albuterol} studies two factors in demonstrations: (1) context length and (2) ratio of positive examples. In \cref{fig:albuterol-contextlen}, longer context includes more molecular examples and supports better ICL performance. This aligns with the rationale of motif-level tokenization, which captures more examples within a fixed context. \cref{fig:albuterol-pos} shows that diverse demonstrations are important for ICL to represent the task accurately, while only positive examples are insufficient. 
This is because positive, medium, and negative examples together provide a holistic view of the task context, and \method pretrained on such contexts is better able to infer latent concepts from diverse examples. In \cref{fig:albuterol-pos}, we also observe that fewer positive examples may still yield reasonable results. We investigate this further in~\cref{subsec:case-study} to assess whether \method can infer positive examples (score > 0.5) using only negative examples with scores below 0.5.

\begin{figure}[ht]
    \centering
    \includegraphics[width=0.8\textwidth]{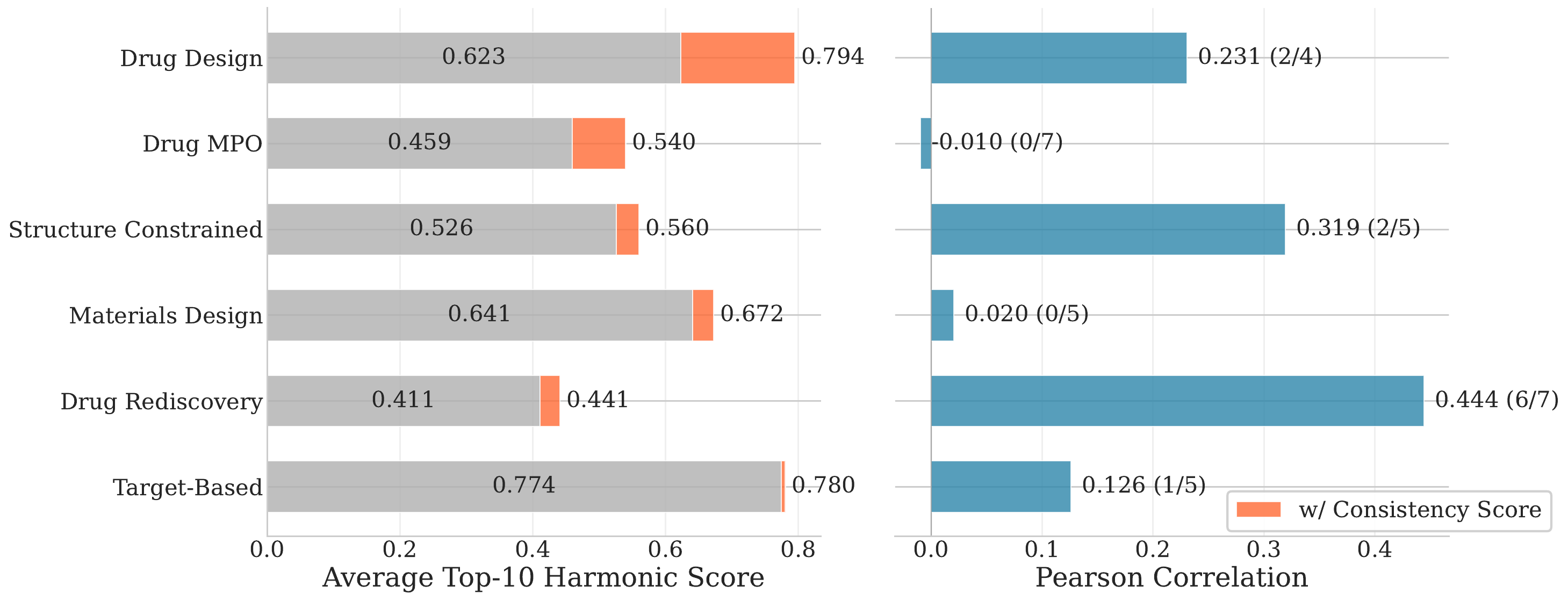}
    \caption{Ablation studies on context consistency scores: (1) left shows improvements; (2) right shows the relationship between consistency score and oracle scores.}
    \label{fig:consistency-for-icl}
\end{figure}

\textbf{ICL with Consistency Scores:}
We ablate consistency scores and analyze their correlation with target scores in~\cref{fig:consistency-for-icl}. Using the consistency score as a confidence filter improves performance across task categories, with gains from 0.8\% to 27.5\%. The second figure shows the correlation between the consistency and target scores. Moderate correlation appears in tasks with explicit structural constraints, such as drug rediscovery and structure-constrained design. For property-driven tasks (drug MPO and materials design), high fingerprint-based consistency with positive examples does not always correlate with high target scores. In these cases, latent concepts may rely on subtle substructures (e.g., methyl groups~\citep{liu2022graph}) that standard fingerprints fail to capture. 
Interestingly, context consistency still improves performance in these tasks. A possible reason is that the score helps filter out false positive generations.

\subsection{Case Studies}\label{subsec:case-study}

\begin{figure}[t]
    \centering
    \includegraphics[width=0.8\textwidth]{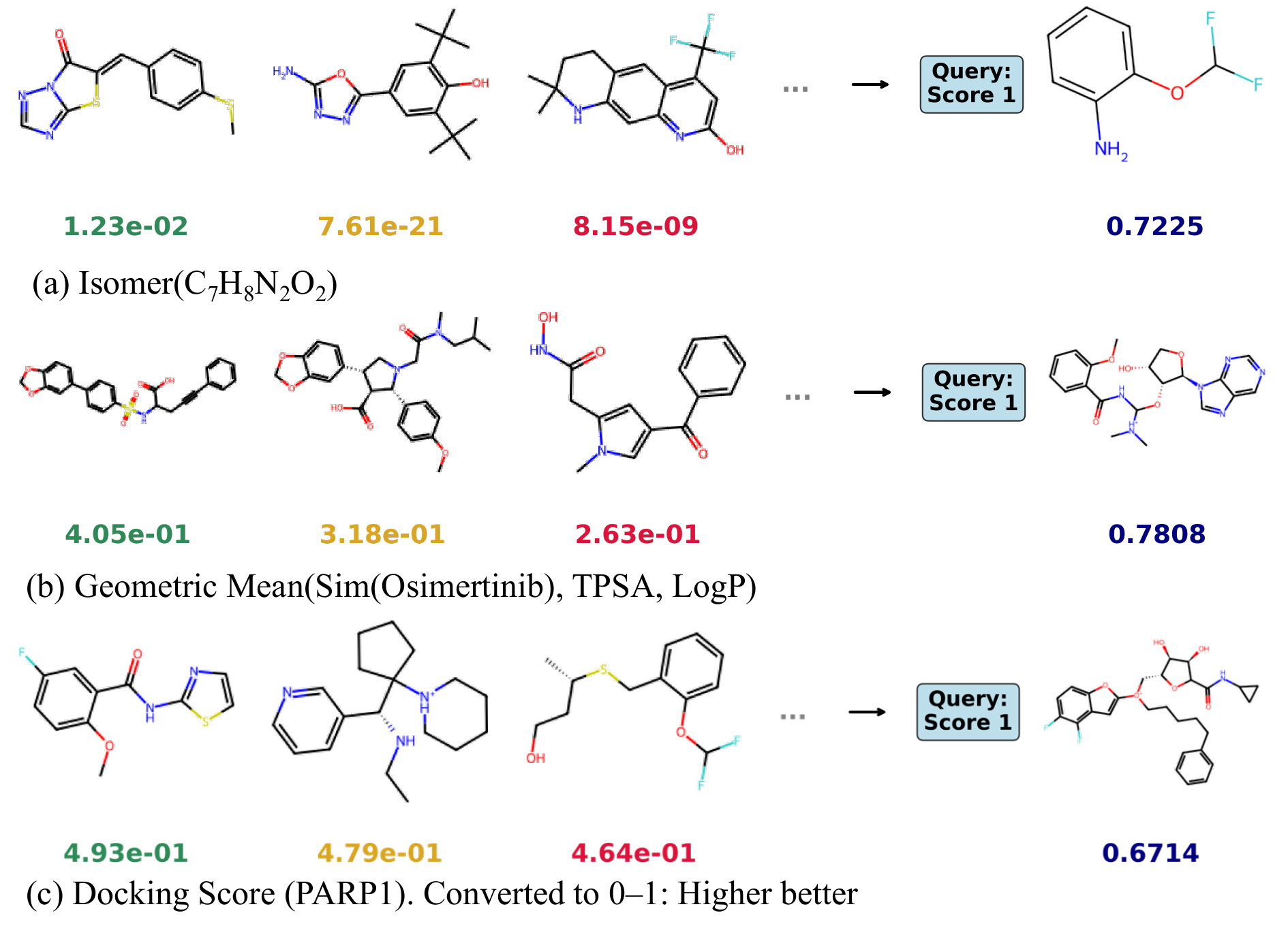}
    \caption{Learning from negative demonstrations (score $<0.5$) to infer a target with score 1. All demonstrations are shown in~\cref{fig:sub-mpo,fig:sub-target,fig:sub-structure}, with only three displayed here.}
    \label{fig:study-nopos}
\end{figure}

\begin{figure}[t]
    \centering
    \includegraphics[width=0.98\textwidth]{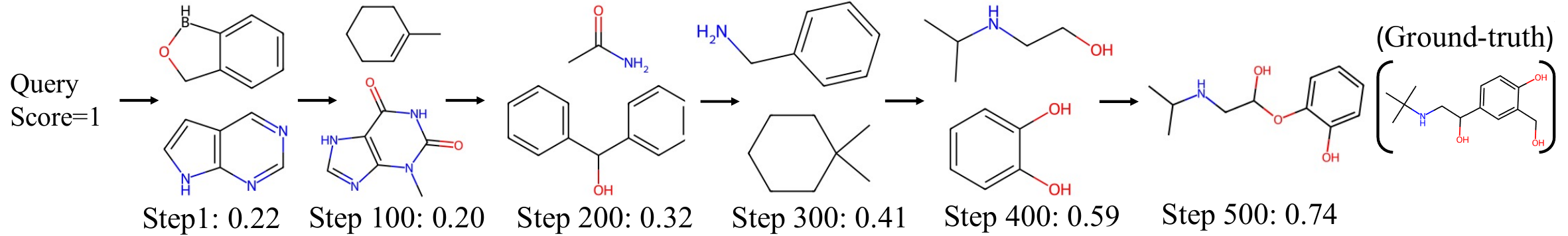}
    \caption{Diffusion trajectory for Albuterol drug rediscovery: we sample five intermediate diffusion steps and score them with the Albuterol Oracle, which computes similarity to the ground truth. 
    }
    \label{fig:study-trajectory}
\end{figure}

\cref{fig:study-nopos,fig:study-trajectory} present two studies for \method.
In extreme cases of inverse molecular design, demonstration sets may contain only negative examples, i.e., all scores $< 0.5$. In~\cref{fig:study-nopos}, we study whether \method can still generate positive candidates when prompted solely with negative examples. \cref{fig:study-nopos} presents the results for (a) structure-constrained design, (b) drug multi-objective optimization (MPO), and (c) target-based design. These findings suggest two insights: (1) negative demonstrations convey informative signals about the task concept, and (2) after pretraining, the posterior over the concept-to-structure mapping allows \method to generate desirable candidates that are aligned with the concept yet structurally distinct from the negative examples.
\cref{fig:study-trajectory} is the generation trajectory from diffusion models. The task score, measured as structural similarity to Albuterol, rises from 0.22 at initial sampling to 0.74. This shows that the diffusion model refines the molecule toward the desired structure step by step with demonstrations.

\section{Related Work}
\textbf{Inverse Molecular Design:}
Molecular optimization uses diverse approaches, including genetic algorithms, Monte Carlo Tree Search~\citep{jensen2019graph}, and Bayesian optimization~\citep{shahriari2015taking}, applied to representations such as fingerprints, SMILES, graphs, and synthetic pathways~\citep{gao2021amortized}. \citet{gao2022sample} benchmarked 25 optimization methods and found that older models, such as genetic algorithms, remain competitive. However, existing benchmarks require on the order of 10,000 oracle calls, which is costly and limits applicability when single calls are expensive.
Deep learning models offer an alternative by modeling the joint distribution of atoms and bonds without Oracle calls. GDSS applies noise and denoising in continuous space for graphs~\citep{jo2022score}. DiGress~\citep{vignac2022digress} introduces discrete noise through transition matrices based on marginal atom and bond distributions. Graph DiTs~\citep{liu2024graph} extend scalable diffusion transformers~\citep{peebles2023scalable} to discrete graphs. Yet, training diffusion models still requires hundreds of labeled molecules and is limited to specific tasks. Recent efforts explore chemical foundation models based on LLMs~\citep{yu2024llasmol,liu2024multimodal}, but their applications are either diverted to other molecular tasks, such as property prediction, or rely on fine-tuning within a restricted scope of design tasks.

\textbf{In-Context Learning:}
ICL is an emergent ability observed in LLMs~\citep{brown2020language,chan2022data}. Empirical and theoretical studies investigate this phenomenon from three perspectives: models, data, and learning mechanisms~\citep{xie2021explanation,min2022rethinking}. For the learning mechanism, ICL can be interpreted as implicit Bayesian inference~\citep{xie2021explanation}, where pretraining data are generated from latent concepts and the posterior distribution marginalizes over them for inference.
On the model side, \citet{garg2022can} trained Transformers from scratch on prompt-style input–label pairs of simple functions and found performance comparable to task-specific algorithms. \citet{bhattamishra2023understanding} compared Transformers with attention-free models and showed they do not match Transformer performance across tasks. 
On the data side, \citet{chan2022data} found that Transformers outperform recurrent models (e.g., LSTMs) on data with distributional properties resembling natural language, such as burstiness (words appearing in clusters) and query tasks with many rare classes.
\citet{singh2025strategy} analyzed the strategy competition between ICL and in-weight learning, showing that the asymptotic strategy depends on in-weight information but is also context-constrained. This aligns with~\citep{chan2022data}, suggesting that a foundation model should support both capacities. A skewed Zipfian distribution over tasks (e.g.,~\cref{fig:task-freq}) balances learning by storing common task information in weights while developing ICL ability from the long tail of rare tasks.

\section{Conclusion}
We presented \method, a demonstration-conditioned diffusion Transformer model for in-context molecular design. We constructed a large-scale pretraining dataset with over one million molecules and 155K unique biological assays and material properties, yielding millions of demonstration–target pairs. Using this dataset, we pretrained a 0.7B-parameter model and showed that it matches or outperforms much larger LLMs and ranks higher than domain- and task-specific methods. To support scalable pretraining, we introduce Node Pair Encoding, a motif-level graph tokenizer that efficiently represents molecules with fewer nodes while preserving reconstruction.
Experiments demonstrate that \method is a promising molecular foundation model, highlighting its potential to scale further with larger models, broader datasets, and greater compute.




\bibliography{reference}

\begin{thebibliography}{35}
\providecommand{\natexlab}[1]{#1}
\providecommand{\url}[1]{\texttt{#1}}
\expandafter\ifx\csname urlstyle\endcsname\relax
  \providecommand{\doi}[1]{doi: #1}\else
  \providecommand{\doi}{doi: \begingroup \urlstyle{rm}\Url}\fi

\bibitem[Achiam et~al.(2023)Achiam, Adler, Agarwal, Ahmad, Akkaya, Aleman, Almeida, Altenschmidt, Altman, Anadkat, et~al.]{achiam2023gpt}
Josh Achiam, Steven Adler, Sandhini Agarwal, Lama Ahmad, Ilge Akkaya, Florencia~Leoni Aleman, Diogo Almeida, Janko Altenschmidt, Sam Altman, Shyamal Anadkat, et~al.
\newblock Gpt-4 technical report.
\newblock \emph{arXiv preprint arXiv:2303.08774}, 2023.

\bibitem[Alhossary et~al.(2015)Alhossary, Handoko, Mu, and Kwoh]{alhossary2015fast}
Amr Alhossary, Stephanus~Daniel Handoko, Yuguang Mu, and Chee-Keong Kwoh.
\newblock Fast, accurate, and reliable molecular docking with quickvina 2.
\newblock \emph{Bioinformatics}, 31\penalty0 (13):\penalty0 2214--2216, 2015.

\bibitem[Bertsch et~al.(2024)Bertsch, Ivgi, Xiao, Alon, Berant, Gormley, and Neubig]{bertsch2024context}
Amanda Bertsch, Maor Ivgi, Emily Xiao, Uri Alon, Jonathan Berant, Matthew~R Gormley, and Graham Neubig.
\newblock In-context learning with long-context models: An in-depth exploration.
\newblock \emph{arXiv preprint arXiv:2405.00200}, 2024.

\bibitem[Bhattamishra et~al.(2023)Bhattamishra, Patel, Blunsom, and Kanade]{bhattamishra2023understanding}
Satwik Bhattamishra, Arkil Patel, Phil Blunsom, and Varun Kanade.
\newblock Understanding in-context learning in transformers and llms by learning to learn discrete functions.
\newblock \emph{arXiv preprint arXiv:2310.03016}, 2023.

\bibitem[Brown et~al.(2019)Brown, Fiscato, Segler, and Vaucher]{brown2019guacamol}
Nathan Brown, Marco Fiscato, Marwin~HS Segler, and Alain~C Vaucher.
\newblock Guacamol: benchmarking models for de novo molecular design.
\newblock \emph{Journal of chemical information and modeling}, 59\penalty0 (3):\penalty0 1096--1108, 2019.

\bibitem[Brown et~al.(2020)Brown, Mann, Ryder, Subbiah, Kaplan, Dhariwal, Neelakantan, Shyam, Sastry, Askell, et~al.]{brown2020language}
Tom Brown, Benjamin Mann, Nick Ryder, Melanie Subbiah, Jared~D Kaplan, Prafulla Dhariwal, Arvind Neelakantan, Pranav Shyam, Girish Sastry, Amanda Askell, et~al.
\newblock Language models are few-shot learners.
\newblock \emph{Advances in neural information processing systems}, 33:\penalty0 1877--1901, 2020.

\bibitem[Chan et~al.(2022)Chan, Santoro, Lampinen, Wang, Singh, Richemond, McClelland, and Hill]{chan2022data}
Stephanie Chan, Adam Santoro, Andrew Lampinen, Jane Wang, Aaditya Singh, Pierre Richemond, James McClelland, and Felix Hill.
\newblock Data distributional properties drive emergent in-context learning in transformers.
\newblock \emph{Advances in neural information processing systems}, 35:\penalty0 18878--18891, 2022.

\bibitem[Degen et~al.(2008)Degen, Wegscheid-Gerlach, Zaliani, and Rarey]{degen2008art}
Jorg Degen, Christof Wegscheid-Gerlach, Andrea Zaliani, and Matthias Rarey.
\newblock On the art of compiling and using'drug-like'chemical fragment spaces.
\newblock \emph{ChemMedChem}, 3\penalty0 (10):\penalty0 1503, 2008.

\bibitem[Durant et~al.(2002)Durant, Leland, Henry, and Nourse]{durant2002reoptimization}
Joseph~L Durant, Burton~A Leland, Douglas~R Henry, and James~G Nourse.
\newblock Reoptimization of mdl keys for use in drug discovery.
\newblock \emph{Journal of chemical information and computer sciences}, 42\penalty0 (6):\penalty0 1273--1280, 2002.

\bibitem[Gao et~al.(2021)Gao, Mercado, and Coley]{gao2021amortized}
Wenhao Gao, Roc{\'\i}o Mercado, and Connor~W Coley.
\newblock Amortized tree generation for bottom-up synthesis planning and synthesizable molecular design.
\newblock \emph{arXiv preprint arXiv:2110.06389}, 2021.

\bibitem[Gao et~al.(2022)Gao, Fu, Sun, and Coley]{gao2022sample}
Wenhao Gao, Tianfan Fu, Jimeng Sun, and Connor Coley.
\newblock Sample efficiency matters: a benchmark for practical molecular optimization.
\newblock \emph{Advances in neural information processing systems}, 35:\penalty0 21342--21357, 2022.

\bibitem[Garg et~al.(2022)Garg, Tsipras, Liang, and Valiant]{garg2022can}
Shivam Garg, Dimitris Tsipras, Percy~S Liang, and Gregory Valiant.
\newblock What can transformers learn in-context? a case study of simple function classes.
\newblock \emph{Advances in neural information processing systems}, 35:\penalty0 30583--30598, 2022.

\bibitem[Jensen(2019)]{jensen2019graph}
Jan~H Jensen.
\newblock A graph-based genetic algorithm and generative model/monte carlo tree search for the exploration of chemical space.
\newblock \emph{Chemical science}, 10\penalty0 (12):\penalty0 3567--3572, 2019.

\bibitem[Jo et~al.(2022)Jo, Lee, and Hwang]{jo2022score}
Jaehyeong Jo, Seul Lee, and Sung~Ju Hwang.
\newblock Score-based generative modeling of graphs via the system of stochastic differential equations.
\newblock In \emph{International conference on machine learning}, pp.\  10362--10383. PMLR, 2022.

\bibitem[Kong et~al.(2022)Kong, Huang, Tan, and Liu]{kong2022molecule}
Xiangzhe Kong, Wenbing Huang, Zhixing Tan, and Yang Liu.
\newblock Molecule generation by principal subgraph mining and assembling.
\newblock \emph{Advances in Neural Information Processing Systems}, 35:\penalty0 2550--2563, 2022.

\bibitem[Kuenneth et~al.(2021)Kuenneth, Rajan, Tran, Chen, Kim, and Ramprasad]{kuenneth2021polymer}
Christopher Kuenneth, Arunkumar~Chitteth Rajan, Huan Tran, Lihua Chen, Chiho Kim, and Rampi Ramprasad.
\newblock Polymer informatics with multi-task learning.
\newblock \emph{Patterns}, 2\penalty0 (4), 2021.

\bibitem[Liu et~al.(2024{\natexlab{a}})Liu, Feng, Xue, Wang, Wu, Lu, Zhao, Deng, Zhang, Ruan, et~al.]{liu2024deepseek}
Aixin Liu, Bei Feng, Bing Xue, Bingxuan Wang, Bochao Wu, Chengda Lu, Chenggang Zhao, Chengqi Deng, Chenyu Zhang, Chong Ruan, et~al.
\newblock Deepseek-v3 technical report.
\newblock \emph{arXiv preprint arXiv:2412.19437}, 2024{\natexlab{a}}.

\bibitem[Liu et~al.(2022)Liu, Zhao, Xu, Luo, and Jiang]{liu2022graph}
Gang Liu, Tong Zhao, Jiaxin Xu, Tengfei Luo, and Meng Jiang.
\newblock Graph rationalization with environment-based augmentations.
\newblock In \emph{Proceedings of the 28th ACM SIGKDD Conference on Knowledge Discovery and Data Mining}, pp.\  1069--1078, 2022.

\bibitem[Liu et~al.(2024{\natexlab{b}})Liu, Sun, Matusik, Jiang, and Chen]{liu2024multimodal}
Gang Liu, Michael Sun, Wojciech Matusik, Meng Jiang, and Jie Chen.
\newblock Multimodal large language models for inverse molecular design with retrosynthetic planning.
\newblock \emph{arXiv preprint arXiv:2410.04223}, 2024{\natexlab{b}}.

\bibitem[Liu et~al.(2024{\natexlab{c}})Liu, Xu, Luo, and Jiang]{liu2024graph}
Gang Liu, Jiaxin Xu, Tengfei Luo, and Meng Jiang.
\newblock Graph diffusion transformers for multi-conditional molecular generation.
\newblock \emph{Advances in Neural Information Processing Systems}, 37:\penalty0 8065--8092, 2024{\natexlab{c}}.

\bibitem[Min et~al.(2022)Min, Lyu, Holtzman, Artetxe, Lewis, Hajishirzi, and Zettlemoyer]{min2022rethinking}
Sewon Min, Xinxi Lyu, Ari Holtzman, Mikel Artetxe, Mike Lewis, Hannaneh Hajishirzi, and Luke Zettlemoyer.
\newblock Rethinking the role of demonstrations: What makes in-context learning work?
\newblock \emph{arXiv preprint arXiv:2202.12837}, 2022.

\bibitem[Otsuka et~al.(2011)Otsuka, Kuwajima, Hosoya, Xu, and Yamazaki]{otsuka2011polyinfo}
Shingo Otsuka, Isao Kuwajima, Junko Hosoya, Yibin Xu, and Masayoshi Yamazaki.
\newblock Polyinfo: Polymer database for polymeric materials design.
\newblock In \emph{2011 International Conference on Emerging Intelligent Data and Web Technologies}, pp.\  22--29. IEEE, 2011.

\bibitem[Peebles \& Xie(2023)Peebles and Xie]{peebles2023scalable}
William Peebles and Saining Xie.
\newblock Scalable diffusion models with transformers.
\newblock In \emph{Proceedings of the IEEE/CVF international conference on computer vision}, pp.\  4195--4205, 2023.

\bibitem[Robeson(2008)]{robeson2008upper}
Lloyd~M Robeson.
\newblock The upper bound revisited.
\newblock \emph{Journal of membrane science}, 320\penalty0 (1-2):\penalty0 390--400, 2008.

\bibitem[Shahriari et~al.(2015)Shahriari, Swersky, Wang, Adams, and De~Freitas]{shahriari2015taking}
Bobak Shahriari, Kevin Swersky, Ziyu Wang, Ryan~P Adams, and Nando De~Freitas.
\newblock Taking the human out of the loop: A review of bayesian optimization.
\newblock \emph{Proceedings of the IEEE}, 104\penalty0 (1):\penalty0 148--175, 2015.

\bibitem[Singh et~al.(2025)Singh, Moskovitz, Dragutinovi{\'c}, Hill, Chan, and Saxe]{singh2025strategy}
Aaditya~K Singh, Ted Moskovitz, Sara Dragutinovi{\'c}, Felix Hill, Stephanie~C.Y. Chan, and Andrew~M Saxe.
\newblock Strategy coopetition explains the emergence and transience of in-context learning.
\newblock In \emph{Forty-second International Conference on Machine Learning}, 2025.
\newblock URL \url{https://openreview.net/forum?id=esBoQFmD7v}.

\bibitem[Su et~al.(2024)Su, Ahmed, Lu, Pan, Bo, and Liu]{su2024roformer}
Jianlin Su, Murtadha Ahmed, Yu~Lu, Shengfeng Pan, Wen Bo, and Yunfeng Liu.
\newblock Roformer: Enhanced transformer with rotary position embedding.
\newblock \emph{Neurocomputing}, 568:\penalty0 127063, 2024.

\bibitem[Sun et~al.(2024)Sun, Guo, Yuan, Thost, Owens, Grosz, Selvan, Zhou, Mohiuddin, Pedretti, et~al.]{sun2024representing}
Michael Sun, Minghao Guo, Weize Yuan, Veronika Thost, Crystal~Elaine Owens, Aristotle~Franklin Grosz, Sharvaa Selvan, Katelyn Zhou, Hassan Mohiuddin, Benjamin~J Pedretti, et~al.
\newblock Representing molecules as random walks over interpretable grammars.
\newblock \emph{arXiv preprint arXiv:2403.08147}, 2024.

\bibitem[Sun et~al.(2025)Sun, Yuan, Liu, Matusik, and Chen]{sun2025foundation}
Michael Sun, Weize Yuan, Gang Liu, Wojciech Matusik, and Jie Chen.
\newblock Foundation molecular grammar: Multi-modal foundation models induce interpretable molecular graph languages.
\newblock \emph{arXiv preprint arXiv:2505.22948}, 2025.

\bibitem[Thornton et~al.(2012)Thornton, Robeson, Freeman, and Uhlmann]{thornton2012polymer}
A~Thornton, L~Robeson, B~Freeman, and D~Uhlmann.
\newblock Polymer gas separation membrane database, 2012.
\newblock URL \url{https://research.csiro.au/virtualscreening/membrane-database-polymer-gas-separation-membranes/}.

\bibitem[Vignac et~al.(2022)Vignac, Krawczuk, Siraudin, Wang, Cevher, and Frossard]{vignac2022digress}
Clement Vignac, Igor Krawczuk, Antoine Siraudin, Bohan Wang, Volkan Cevher, and Pascal Frossard.
\newblock Digress: Discrete denoising diffusion for graph generation.
\newblock \emph{arXiv preprint arXiv:2209.14734}, 2022.

\bibitem[Xie et~al.(2021)Xie, Raghunathan, Liang, and Ma]{xie2021explanation}
Sang~Michael Xie, Aditi Raghunathan, Percy Liang, and Tengyu Ma.
\newblock An explanation of in-context learning as implicit bayesian inference.
\newblock \emph{arXiv preprint arXiv:2111.02080}, 2021.

\bibitem[Yang et~al.(2025)Yang, Yu, Li, Liu, Huang, Huang, Jiang, Tu, Zhang, Zhou, et~al.]{yang2025qwen2}
An~Yang, Bowen Yu, Chengyuan Li, Dayiheng Liu, Fei Huang, Haoyan Huang, Jiandong Jiang, Jianhong Tu, Jianwei Zhang, Jingren Zhou, et~al.
\newblock Qwen2. 5-1m technical report.
\newblock \emph{arXiv preprint arXiv:2501.15383}, 2025.

\bibitem[Yu et~al.(2024)Yu, Baker, Chen, Ning, and Sun]{yu2024llasmol}
Botao Yu, Frazier~N Baker, Ziqi Chen, Xia Ning, and Huan Sun.
\newblock Llasmol: Advancing large language models for chemistry with a large-scale, comprehensive, high-quality instruction tuning dataset.
\newblock \emph{arXiv preprint arXiv:2402.09391}, 2024.

\bibitem[Zdrazil et~al.(2024)Zdrazil, Felix, Hunter, Manners, Blackshaw, Corbett, De~Veij, Ioannidis, Lopez, Mosquera, et~al.]{zdrazil2024chembl}
Barbara Zdrazil, Eloy Felix, Fiona Hunter, Emma~J Manners, James Blackshaw, Sybilla Corbett, Marleen De~Veij, Harris Ioannidis, David~Mendez Lopez, Juan~F Mosquera, et~al.
\newblock The chembl database in 2023: a drug discovery platform spanning multiple bioactivity data types and time periods.
\newblock \emph{Nucleic acids research}, 52\penalty0 (D1):\penalty0 D1180--D1192, 2024.

\end{thebibliography}
\bibliographystyle{iclr2026_conference}

\appendix
\section{Details on \method}~\label{addsec:model-details}

\subsection{Graph-level Tokens and Pretraining Loss}

Graph DiTs define a graph-level token that concatenates the node feature with all related edge features. In molecular generation, we use a special type of edge, the null edge, to represent that there is no edge between two nodes. Thus,
The feature vector $\mathbf{x}$ (or $\mathbf{x}^0$) of a graph-level token 
$x=\{M, \{e_{j}\}_{j=1}^d\}$ consists of three components: 
$F_\mathrm{motif}$ motif types, $F_\mathrm{bond}$ bond types, 
and $F_\mathrm{attach}$ attachment specifications. 
Here, $F_\mathrm{motif}$ is the size of the motif vocabulary $\mathcal{M}$, 
$F_\mathrm{bond}=4$ represents null, single, double, and triple bonds, 
and $F_\mathrm{attach}=\arg\max_{M\in\mathcal{M}} |M|$ is the maximum number of atoms in $\mathcal{M}$. \cref{eq-icl-pretrain} can be decomposed as 
$\mathcal{L}_{\mathrm{pretrain}}
= \mathcal{L}_{\mathrm{motif}}
+ \mathcal{L}_{\mathrm{bond}}
+ \mathcal{L}_{\mathrm{attach}}$. Specifically,
\begin{align}\label{eq-pretrain-decomposed}
\mathcal{L}_{\text{pretrain}}
= \mathbb{E}_{q(\mathbf{x})} \, \mathbb{E}_{q(\mathbf{x}^t \mid \mathbf{x})} \Bigg[
& \underbrace{- \log p_{\theta}^{\mathrm{motif}}\!\big( \mathbf{m} \mid \mathbf{x}^t, \mathcal{C}, Y \big)}_{\mathcal{L}_{\mathrm{motif}}} \nonumber \\
& \underbrace{- \sum_{j=1}^{d} \log p_{\theta}^{\mathrm{bond}}\!\big( \mathbf{b}_{j} \mid \mathbf{x}^t, \mathcal{C}, Y \big)}_{\mathcal{L}_{\mathrm{bond}}} \nonumber \\
& \underbrace{- \sum_{j=1}^{d} \log p_{\theta}^{\mathrm{attach}}\!\big( \mathbf{a}_{j} \mid \mathbf{x}^t, \mathcal{C}, Y \big)}_{\mathcal{L}_{\mathrm{attach}}} \Bigg].
\end{align}
To align feature dimensions across tokens, we use the dense edge representation by treating all non-connections as null bonds. 
The resulting feature dimension is $
F = F_\mathrm{motif} + n \times F_\mathrm{bond} + n \times F_\mathrm{attach}, $
where $n$ is the maximum number of nodes in the motif-represented dataset. 
For optimization with~\cref{eq-pretrain-decomposed}, we include the null bond type in $\mathcal{L}_\mathrm{bond}$ but exclude attachment specifications of null edges in $\mathcal{L}_\mathrm{attach}$. 

\subsection{Transition Matrices in Diffusion Models}

We define the transition matrix $\mathbf{Q}$ that perturbs molecules at the motif level to pretrain \method.
We model the joint distribution of nodes and edges with the transition matrix. It is constructed from four submatrices $\mathbf{Q}_V, \mathbf{Q}_{EV}, \mathbf{Q}_E, \mathbf{Q}_{VE}$, 
denoting transitions $\text{node}\to\text{node}$, $\text{edge}\to\text{node}$, 
$\text{edge}\to\text{edge}$, and $\text{node}\to\text{edge}$, respectively:
\begin{equation}\label{eq:graph-transition-def}
    \mathbf{Q}_G = 
    \begin{bmatrix}
    \mathbf{Q}_V &  \mathbf{1}_N^{\prime} \otimes \mathbf{Q}_{VE} \\
    \mathbf{1}_n \otimes \mathbf{Q}_{EV} & \mathbf{1}_{n\times n} \otimes \mathbf{Q}_E
    \end{bmatrix},
\end{equation}
where $\otimes$ denotes the Kronecker product, and $\mathbf{1}_N$, $\mathbf{1}_n^\prime$, 
and $\mathbf{1}_{n\times n}$ are the column vector, row vector, and all-ones matrix, respectively. 
Here $n$ is the number of nodes. For edges, diffusion is applied to bond types only, 
while attachment attributes are optimized and predicted directly by the denoising model 
as in~\cref{eq-pretrain-decomposed}. For categorical sampling, we separate the unnormalized logits of node and edge from the model outputs, compute probabilities for each motif and bond individually before sampling.  

To obtain the transition matrices, we use the prior from the pretraining data. 
The noisy distribution is defined as the marginal distributions of motif types $\mathbf{m}_{V}$ 
and bond types $\mathbf{m}_{E}$. The transition matrices are defined as  
$\mathbf{Q}_V = \bar{\alpha}^t \mathbf{I} + (1-\bar{\alpha}^t)\mathbf{1}\mathbf{m}_{V}^\prime$ 
and  
$\mathbf{Q}_E = \bar{\alpha}^t \mathbf{I} + (1-\bar{\alpha}^t)\mathbf{1}\mathbf{m}_{E}^\prime$, 
where $\mathbf{m}^\prime$ denotes the transpose and $\mathbf{I}$ is the identity matrix.
We compute co-occurrence frequencies of motif and bond types in training graphs 
to obtain the marginal distributions $\mathbf{m}_{EV}$ and $\mathbf{m}_{VE}$. 
Each row in $\mathbf{m}_{EV}$ gives the probability of co-occurring motifs for a bond type, 
and $\mathbf{m}_{VE}$ is its transpose. The transition matrices are then defined as  
$\mathbf{Q}_{EV} = \bar{\alpha}^t \mathbf{I} + (1-\bar{\alpha}^t)\mathbf{1}\mathbf{m}_{EV}^\prime$  
and  
$\mathbf{Q}_{VE} = \bar{\alpha}^t \mathbf{I} + (1-\bar{\alpha}^t)\mathbf{1}\mathbf{m}_{VE}^\prime$, where $\bar{\alpha}^t$ is cumulative noise coefficient in diffusion. The cosine schedule is chosen as $\bar{\alpha}^t = \operatorname{cos}(0.5 \pi (t/T +s) / (1+s) )^2$.

\subsection{Details on Consistency Score}\label{addsec:consistency-score}

Given a query $Y$, demonstrations $\mathcal{C}_i = \{(X_{ij}, Y_{ij})\}_{j=1}^{L}$ are divided into positive $\mathcal{C}^{\mathrm{pos}}$, medium $\mathcal{C}^{\mathrm{med}}$, and negative $\mathcal{C}^{\mathrm{neg}}$ examples to guide ICL. For a generated molecule $X$, we use the Tanimoto similarity of fingerprints as the similarity measure. We compute the similarity between $X$ and all molecules in each group and average them to obtain group-wise similarity scores
\[
\mathrm{sim}^{\mathrm{pos}}, \; \mathrm{sim}^{\mathrm{med}}, \; \mathrm{sim}^{\mathrm{neg}} \in [0,1].
\]

\textbf{Difference-based score.}  
We compute margin differences between groups:
\[
d_{\mathrm{pos,med}} = \max(\mathrm{sim}^{\mathrm{pos}} - \mathrm{sim}^{\mathrm{med}}, 0), \quad
d_{\mathrm{med,neg}} = \max(\mathrm{sim}^{\mathrm{med}} - \mathrm{sim}^{\mathrm{neg}}, 0),
\]
\[
d_{\mathrm{pos,neg}} = \max(\mathrm{sim}^{\mathrm{pos}} - \mathrm{sim}^{\mathrm{neg}}, 0).
\]
The normalized difference-based score is
\[
s_{\mathrm{diff}} = \min\left( \frac{d_{\mathrm{pos,med}} + d_{\mathrm{med,neg}} + d_{\mathrm{pos,neg}}}{3}, \, 1 \right).
\]

In experiments, the consistency score can be computed efficiently before applying Oracle functions. For example, we generate 1000 molecules and select the top 100 with the highest consistency scores. These molecules better follow the order of structural similarity across positive, medium, and negative examples. This removes poor generations that conflict with the demonstration semantics and increases confidence that selected molecules align with the query scores. \cref{tab:consistency-improve} reports empirical improvements across task categories, each containing 4–7 tasks (\cref{addsec:setups}).

\begin{table}[ht]
\centering
\caption{Improvement with the consistency score (average Top-10 harmonic scores).}
\label{tab:consistency-improve}
\begin{tabular}{lccc}
\toprule
Category & Without & With & Improvement \\
\midrule
Drug Design            & 0.6230 & 0.7943 & $+27.5\%$ \\
Drug MPO               & 0.4592 & 0.5400 & $+17.6\%$ \\
Drug Rediscovery       & 0.4110 & 0.4407 & $+7.2\%$  \\
Structure Constrained  & 0.5258 & 0.5598 & $+6.5\%$  \\
Materials Design       & 0.6407 & 0.6724 & $+4.9\%$  \\
Target-Based           & 0.7745 & 0.7803 & $+0.8\%$  \\
\bottomrule
\end{tabular}
\end{table}

\section{Details on Pretraining}~\label{addsec:pretraining-details}

The final pretraining dataset contains 1,084,566 molecules (polymers) and 155,150 unique assays or properties, yielding 1,639,515 tasks. These are constructed from ChEMBL~\citep{zdrazil2024chembl} and multiple polymer data sources~\citep{otsuka2011polyinfo,thornton2012polymer,kuenneth2021polymer}. Each task has a query molecule–score pair with the query score fixed at 1. Up to 45 molecules are grouped into positive, medium, and negative demonstrations based on their scores. The query molecule is the target, while the query score and demonstrations serve as inputs to \method during pretraining on~\cref{eq-icl-pretrain}.
For pretraining with a fixed maximum context window, we allocate half the window to positive demonstrations and one quarter each to medium and negative demonstrations, after excluding the target molecule.

\subsection{Pretraining dataset}\label{addsec:pretraining-data-intro}

\paragraph{ChEMBL dataset}
We constructed molecular activity contexts from the ChEMBL database (version 35), which provides a large collection of bioactivity measurements across diverse assays. ChEMBL standardizes published activity types, values, and units into a unified variable, $p\mathrm{ChEMBL} = -\log(\text{molar IC}{50}, \text{XC}{50}, \text{EC}{50}, \text{AC}{50}, K_i, K_d, \text{or Potency})$. This value places different measures of half-maximal response, potency, or affinity on a comparable negative logarithmic scale. For example, an IC\text{50} of 1 nanomolar ($1\times 10^{-9}$ M) corresponds to a $p\mathrm{ChEMBL}$ value of 9.
We extracted assay-level activity values ($p\mathrm{ChEMBL}$).
For each assay, molecules were grouped according to their recorded activities. Within each group, we selected anchor molecules with strong activity ($p\mathrm{ChEMBL} > 6$) as targets for building demonstrations. Each anchor was compared against all other molecules in the same assay to compute normalized distances, defined as the relative difference between the anchor’s $p\mathrm{ChEMBL}$ value and that of the candidate molecule, converted to the range $[0,1]$. Specifically, for an anchor with value $v_a$ and a candidate with value $v_c$, the normalized distance was given by $d = (v_1 - v_c)/10. $
Based on this distance, we partitioned candidate molecules into three categories relative to the anchor. Molecules with distances in $[0, 0.25)$ correspond to candidates with activity between $75\%$ and $100\%$ of the anchor and were assigned to the positive context. Molecules with distances in $[0.25, 0.5)$ correspond to candidates with activity between $50\%$ and $75\%$ of the anchor and were assigned to the medium context. Molecules with distances $[0.5, 1.0]$ correspond to candidates with activity below $50\%$ of the anchor and were assigned to the negative context. From each category, we sampled up to $15$ molecules to balance neighborhood size. Thus, there are up to 45 demonstration molecules for each task. Not all of them are used during pretraining due to the constraint of maximum context length.
This procedure produced triplets of anchor molecules and their associated positive, medium, and negative contexts. 

\paragraph{Polymeric materials}
We have polymeric material datasets from different sources, including PolyInfo~\citep{otsuka2011polyinfo}, MSA~\citep{thornton2012polymer}, and from~\citep{kuenneth2021polymer}. 
We considered a wide range of polymer properties spanning several categories, including thermal properties (e.g., heat capacity, glass transition temperature, melting temperature, and thermal conductivity), electronic properties (e.g., ionization energy, electron affinity, and band gap), structural properties (e.g., density, crystallinity, and radius of gyration), and transport properties (e.g., gas diffusion, solubility, and permeability coefficients).
For each property, raw values were normalized to the unit interval using min–max scaling, with logarithmic transformation applied when dynamic ranges exceeded $1000$ and non-negative shifts applied when necessary. Each polymer with valid property values was treated as an anchor, and pairwise distances in normalized property space were computed against all other polymers. Candidate molecules were partitioned into positive $[0, 0.25)$, medium $[0.25, 0.5)$, and negative $[0.5, 1]$ contexts, with up to $15$ examples sampled per category based on smallest distances.

\subsection{Tokenizer Preparation}\label{addsec:tokenizer-prepare}

\begin{algorithm}[ht]
\caption{Node Pair Encoding (\token) with Constraints}
\label{alg:npe-merge}
\begin{algorithmic}[1]
\REQUIRE molecule list $\mathcal{D}$, motif vocabulary $\mathcal{M} = \emptyset$, max size $K$, ring count threshold $N_\text{ring}$
\ENSURE motif vocabulary $\mathcal{M}$
\STATE Initialize each molecule $X \in \mathcal{D}$ with atom-level and ring-based motifs
\STATE Count frequencies of ring-based motifs across $\mathcal{D}$
\STATE Add all periodic-table elements, polymerization ``*'', and top-$N_\text{ring}$ frequent rings to $\mathcal{M}$
\WHILE{$|\mathcal{M}| < K$}
    \STATE \textbf{(1) Merge Neighbor:} Initialize empty multiset $\mathcal{S} \leftarrow \emptyset$
    \FOR{each molecule $X \in \mathcal{D}$}
        \FOR{each motif $m$ from $X$}
            \FOR{each adjacent motif $m'$ in $X$ such that $m$ and $m'$ are mergeable under structural constraints (e.g., rings treated as units)}
                \STATE form new motif $m \leftarrow m \cup m'$
                \STATE add $m$ to multiset $\mathcal{S}$ with frequency count
            \ENDFOR
        \ENDFOR
    \ENDFOR
    \STATE \textbf{(2) Frequency Selection:} Find most frequent motif $m^* \in \mathcal{S}$
    \STATE \textbf{(3) Update Graph:}
    \FOR{each molecule $X \in \mathcal{D}$}
        \FOR{each pair of adjacent motifs $(m, m')$ in $X$}
            \IF{their merged form equals $m^*$}
                \STATE replace $m$ and $m'$ with $m^*$ in $X$
            \ENDIF
        \ENDFOR
    \ENDFOR
    \STATE Add $m^*$ to $\mathcal{M}$ if not already in it
\ENDWHILE
\RETURN $\mathcal{M}$
\end{algorithmic}
\end{algorithm}

\begin{figure}[ht]
    \centering
    \includegraphics[width=0.95\textwidth]{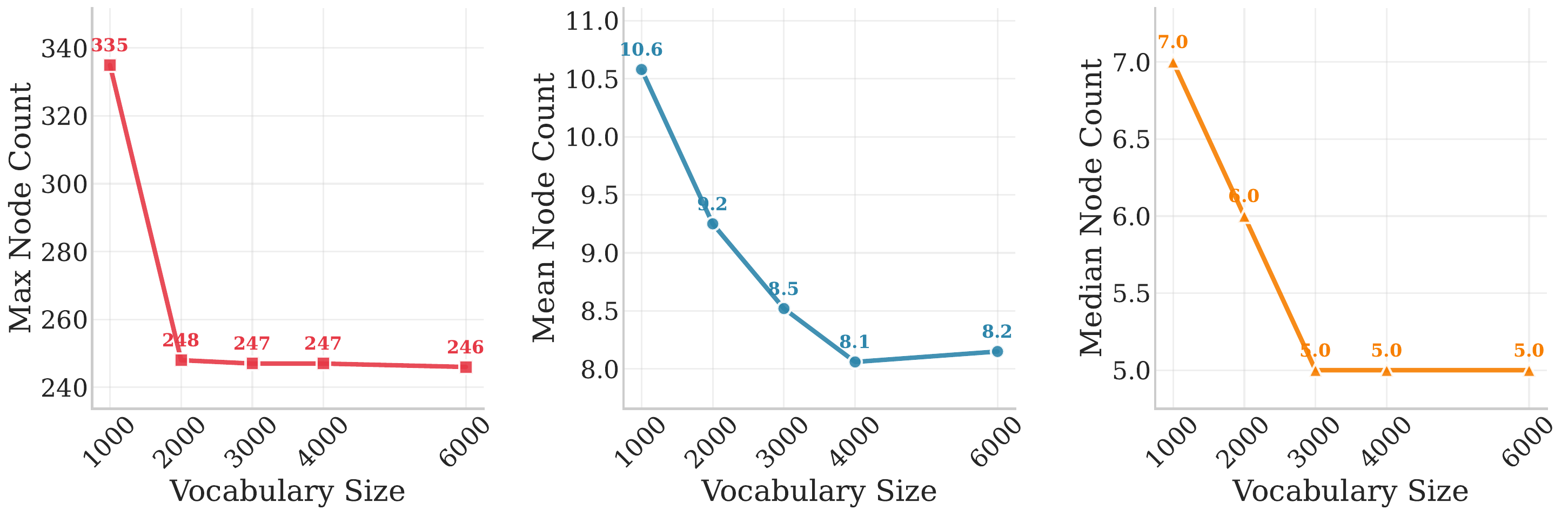}
    \caption{Change in node count with varying motif vocabulary size.}
    \label{fig:motif-voc-size-vs-node}
\end{figure}

We present \token in~\cref{alg:npe-merge}, inspired by both the classic BPE and~\citep{kong2022molecule}. We build the tokenizer with \token on the pretraining data. To choose the motif vocabulary size, we analyze the number of nodes in motif-represented molecular graphs as the vocabulary size varies (\cref{fig:motif-voc-size-vs-node}). We report mean, max, and median counts. We set $K_\text{ring} = K/10$, except for $K=6000$, where $K_\text{ring}=300$. When $K \geq 3000$, the mean and max node counts no longer significantly decrease, and the median remains unchanged. Therefore, we select $K=3000$ with $K_\text{ring}=300$ for pretraining.

\begin{figure}[ht]
    \centering
    \includegraphics[width=0.8\textwidth]{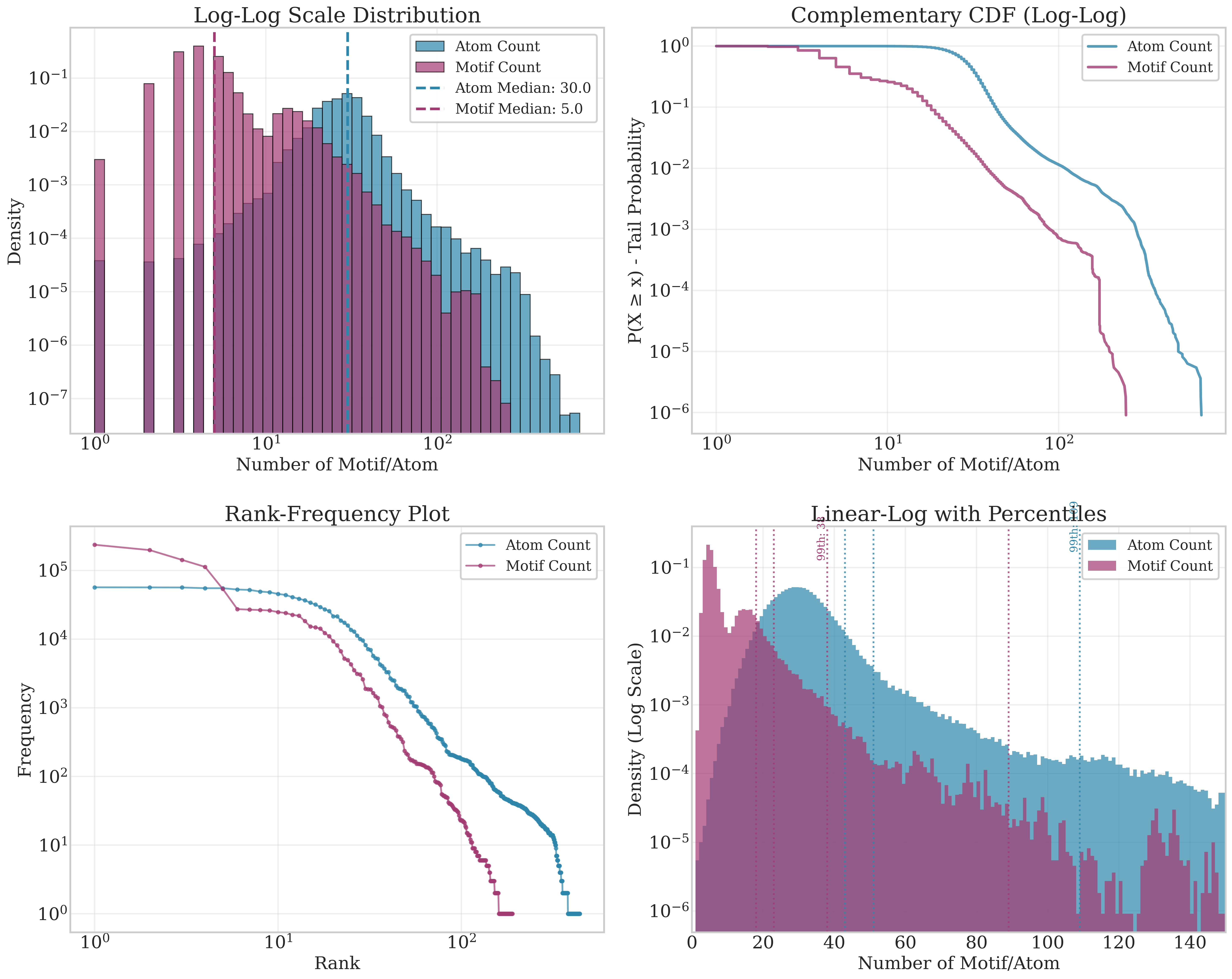}
    \caption{Comparison of the number of nodes in atom- and motif-based representations for 1 million molecules from the pretraining set.}
    \label{fig:num-node-full}
\end{figure}

Next, \cref{fig:num-node-full} presents the tokenization results on the pretraining dataset. log-scale distributions of motif- and atom-level node counts. Both representations exhibit heavy-tailed behavior, as shown by the rank-frequency plots and complementary cumulative distribution functions (CCDFs).

\begin{figure}[ht]
    \centering
    \includegraphics[width=0.8\textwidth]{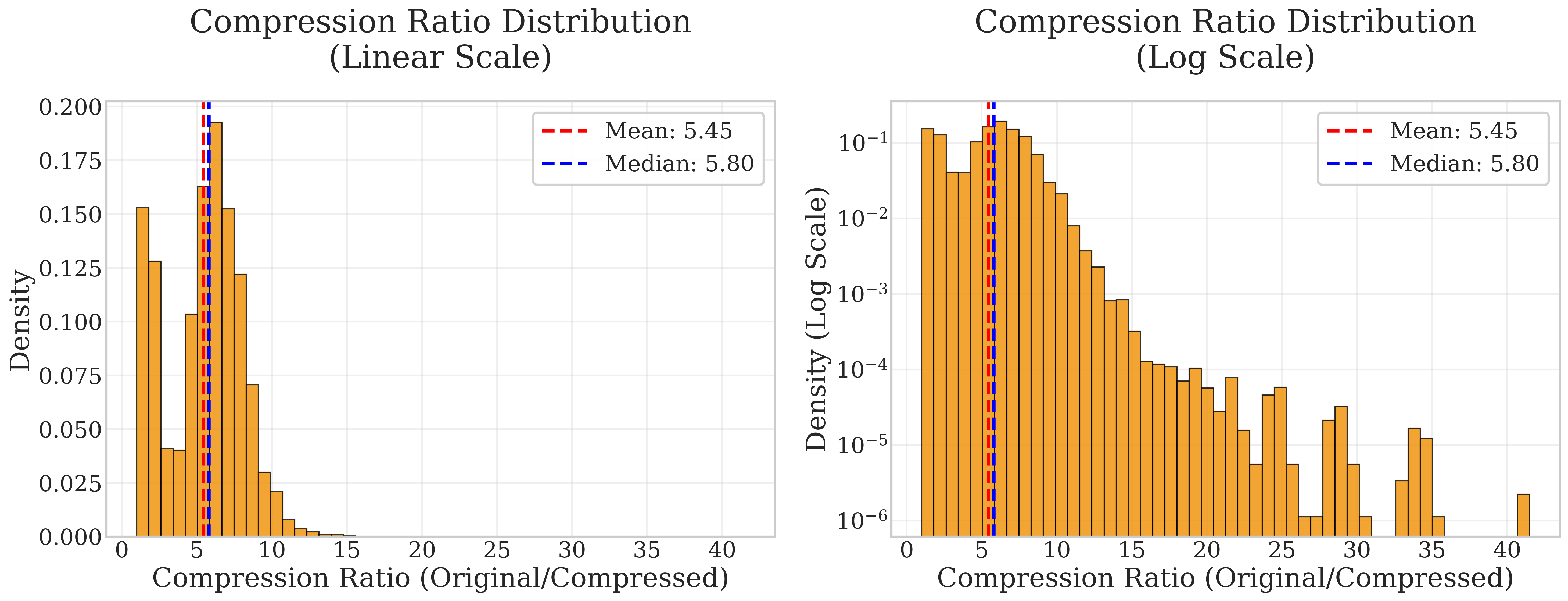}
    \caption{Analysis on the compression ratio.}
    \label{fig:compress-ratio}
\end{figure}

\cref{fig:compress-ratio} shows the distribution of compression ratios, defined as $\frac{\text{Uncompressed Graph Size}}{\text{Compressed Graph Size}}$, in both linear and logarithmic scale. The ratio ranges from 1 to 40, with a median and mean around 5.5, indicating a consistent reduction in graph size by approximately a factor of five.

\begin{figure}[ht]
    \centering
    \includegraphics[width=0.8\textwidth]{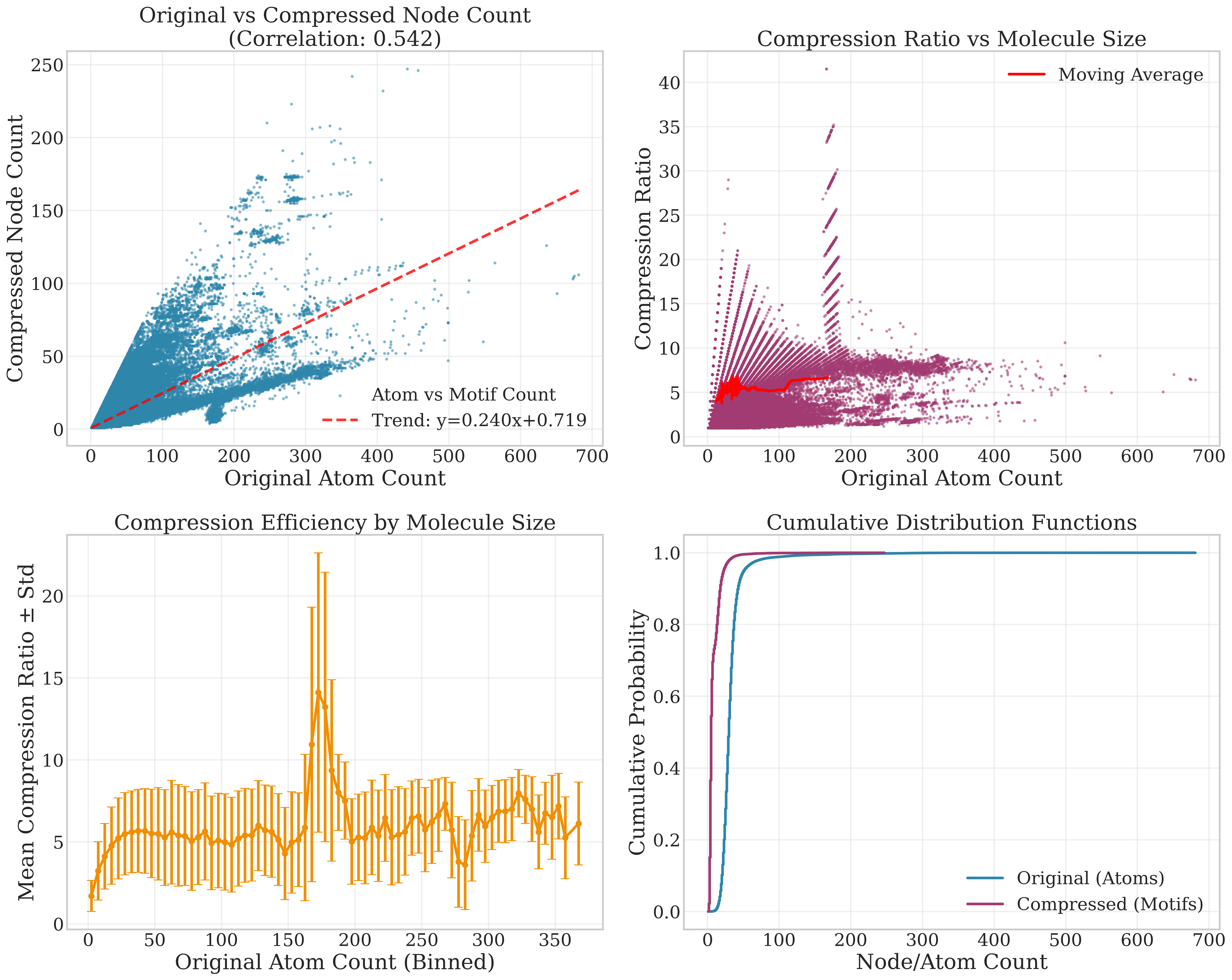}
    \caption{Analysis on the relationship between atom- and motif-level representations.}
    \label{fig:compress-advanced}
\end{figure}

\cref{fig:compress-advanced} provides a detailed analysis of the relationship between atom-level and motif-level representations. We observe a mild positive correlation: larger molecules tend to yield higher compression ratios. Notably, molecules with 150 to 200 atoms are reduced by up to a factor of 15, demonstrating efficient compression at larger scales.

\subsection{Model Pretraining}

\begin{figure*}[t]
    \centering
    \begin{subfigure}{1\textwidth}
        \centering
        \includegraphics[width=\linewidth]{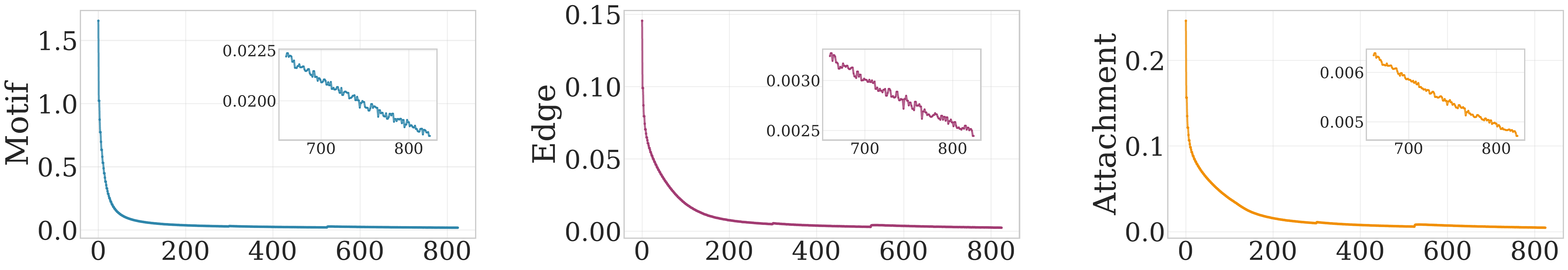}
        \caption{Training losses over steps}
        \label{fig:training-losses}
    \end{subfigure}
    \begin{subfigure}{1\textwidth}
        \centering
        \includegraphics[width=\linewidth]{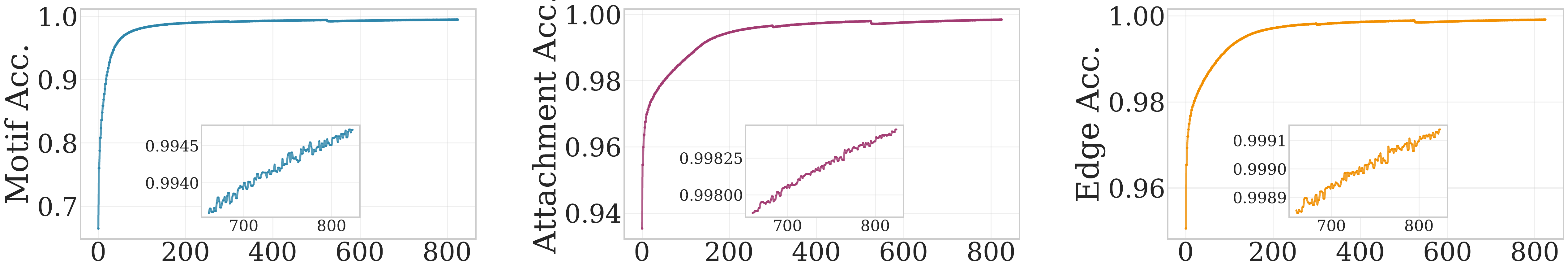}
        \caption{Training accuracies over steps}
        \label{fig:training-accuracies}
    \end{subfigure}
    \caption{Training curves showing (a) losses and (b) accuracies.}
    \label{fig:pretrain-curve}
\end{figure*}

\begin{figure*}[t]
    \centering
    \includegraphics[width=0.8\linewidth]{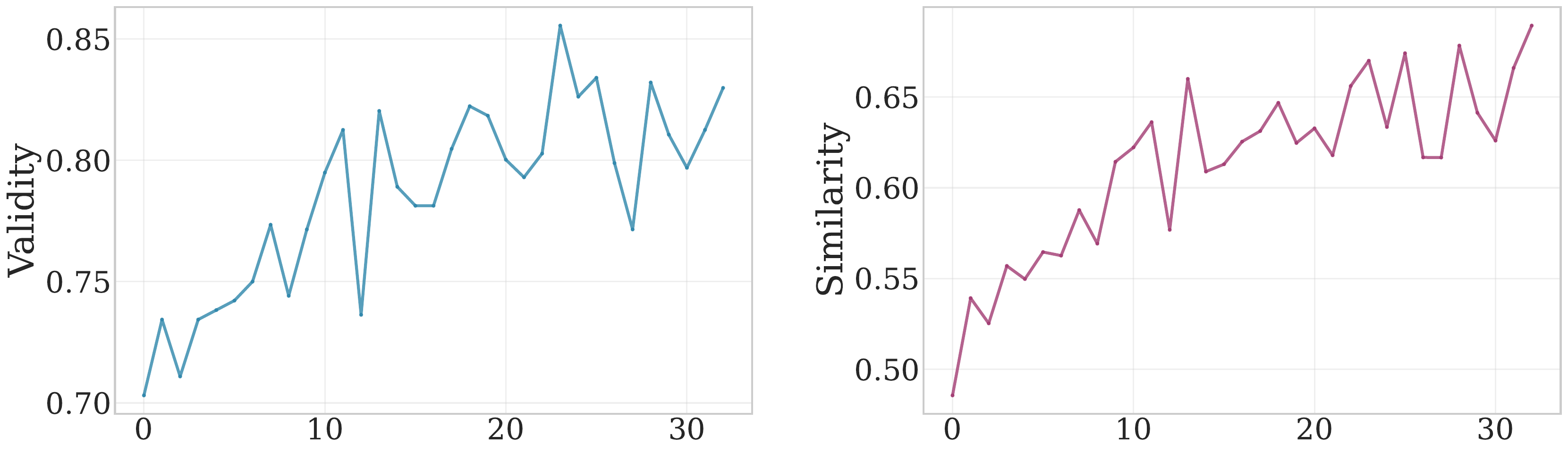}
    \caption{Generation validity and structure similarity to the target during pretraining.}
    \label{fig:pretrain-sample-curve}
\end{figure*}

We pretrain a 0.7B-parameter model (Transformer depth 24, hidden size 1280, 16 heads, MLP ratio 4) for 550 epochs, requiring 49 days on 2–4 H100 GPUs, or about 146 GPU days. We monitor training loss and reconstruction accuracy for each component in~\cref{eq-pretrain-decomposed}. As shown in~\cref{fig:pretrain-curve}, loss decreases and accuracy increases throughout training. Near the end, values plateau but still show incremental gains. Pretraining was stopped once reconstruction accuracy exceeded 0.99 for all components due to resource limits.
During training, we also generated 512 molecules at sampled steps using the validation set. In~\cref{fig:pretrain-sample-curve}, we report chemical validity and structural similarity to ground truth measured by MACCS fingerprints~\citep{durant2002reoptimization}. Both metrics improve with training and reach about 0.83 validity and 0.69 similarity.

These trends in loss, accuracy, validity, and similarity indicate that larger models, more data, and additional compute could further improve \method as a molecular foundation model.

\section{Details on Experiments}

\subsection{Details on Experimental Set-ups}\label{addsec:setups}

\begin{table}[htbp]
\centering
\caption{Benchmark Task Statistics: Example Counts and Score Ranges [min, median, max]}
\label{tab:benchmark_task}
\begin{adjustbox}{width=\textwidth}
\begin{tabular}{l|c|c|c|c|p{2.2cm}|p{2.2cm}|p{2.2cm}|p{2.2cm}}
\toprule
\textbf{Task Name} & \textbf{Total} & \textbf{Pos} & \textbf{Med} & \textbf{Neg} & \textbf{All Scores} & \textbf{Pos Scores} & \textbf{Med Scores} & \textbf{Neg Scores} \\
\midrule
\rowcolor{lightgray} \multicolumn{9}{c}{\textbf{Drug Rediscovery}} \\
Albuterol Similarity & 450 & 150 & 150 & 150 & [0.075, 0.537, 1.000] & [0.752, 0.807, 1.000] & [0.502, 0.537, 0.744] & [0.075, 0.219, 0.476] \\
Celecoxib Rediscovery & 450 & 150 & 150 & 150 & [0.015, 0.548, 1.000] & [0.753, 0.786, 1.000] & [0.505, 0.548, 0.735] & [0.015, 0.153, 0.367] \\
Median 1 & 150 & 0 & 0 & 150 & [0.000, 0.057, 0.419] &  &  & [0.000, 0.057, 0.419] \\
Median 2 & 150 & 0 & 0 & 150 & [0.038, 0.122, 0.413] &  &  & [0.038, 0.122, 0.413] \\
Mestranol Similarity & 450 & 150 & 150 & 150 & [0.004, 0.538, 1.000] & [0.752, 0.824, 1.000] & [0.500, 0.538, 0.713] & [0.004, 0.150, 0.379] \\
Thiothixene Rediscovery & 336 & 36 & 150 & 150 & [0.019, 0.510, 1.000] & [0.753, 0.790, 1.000] & [0.505, 0.560, 0.736] & [0.019, 0.155, 0.354] \\
Troglitazone Rediscovery & 383 & 83 & 150 & 150 & [0.018, 0.526, 1.000] & [0.752, 0.792, 1.000] & [0.504, 0.553, 0.727] & [0.018, 0.143, 0.250] \\
\midrule
\rowcolor{lightgray} \multicolumn{9}{c}{\textbf{Drug MPO}} \\
Amlodipine Mpo & 450 & 150 & 150 & 150 & [0.000, 0.517, 0.871] & [0.750, 0.784, 0.871] & [0.502, 0.517, 0.742] & [0.000, 0.145, 0.500] \\
Fexofenadine Mpo & 450 & 150 & 150 & 150 & [0.000, 0.570, 0.960] & [0.750, 0.772, 0.960] & [0.501, 0.570, 0.720] & [0.000, 0.125, 0.498] \\
Osimertinib Mpo & 450 & 150 & 150 & 150 & [0.000, 0.638, 0.908] & [0.750, 0.761, 0.908] & [0.501, 0.638, 0.747] & [0.000, 0.047, 0.493] \\
Perindopril Mpo & 306 & 6 & 150 & 150 & [0.000, 0.502, 0.790] & [0.766, 0.778, 0.790] & [0.502, 0.522, 0.680] & [0.000, 0.114, 0.415] \\
Ranolazine Mpo & 450 & 150 & 150 & 150 & [0.000, 0.575, 0.867] & [0.750, 0.764, 0.867] & [0.502, 0.575, 0.743] & [0.000, 0.052, 0.454] \\
Sitagliptin Mpo & 389 & 89 & 150 & 150 & [0.000, 0.610, 0.841] & [0.750, 0.768, 0.841] & [0.500, 0.641, 0.748] & [0.000, 0.000, 0.213] \\
Zaleplon Mpo & 300 & 0 & 150 & 150 & [0.000, 0.493, 0.637] &  & [0.501, 0.528, 0.637] & [0.000, 0.003, 0.486] \\
\midrule
\rowcolor{lightgray} \multicolumn{9}{c}{\textbf{Structure Constrained Design}} \\
Deco Hop & 450 & 150 & 150 & 150 & [0.252, 0.529, 0.953] & [0.793, 0.842, 0.953] & [0.502, 0.529, 0.677] & [0.252, 0.283, 0.500] \\
Isomers C7H8N2O2 & 450 & 150 & 150 & 150 & [0.000, 0.592, 1.000] & [0.799, 0.819, 1.000] & [0.535, 0.592, 0.741] & [0.000, 0.000, 0.449] \\
Isomers C9H10N2O2Pf2Cl & 450 & 150 & 150 & 150 & [0.000, 0.561, 0.882] & [0.767, 0.779, 0.882] & [0.503, 0.561, 0.720] & [0.000, 0.000, 0.386] \\
Scaffold Hop & 450 & 150 & 150 & 150 & [0.333, 0.510, 0.828] & [0.754, 0.782, 0.828] & [0.500, 0.510, 0.627] & [0.333, 0.380, 0.444] \\
Valsartan Smarts & 188 & 23 & 15 & 150 & [0.000, 0.000, 0.975] & [0.757, 0.806, 0.975] & [0.512, 0.686, 0.739] & [0.000, 0.000, 0.000] \\
\midrule
\rowcolor{lightgray} \multicolumn{9}{c}{\textbf{Drug Design}} \\
DRD2 & 450 & 150 & 150 & 150 & [0.000, 0.623, 1.000] & [0.760, 0.961, 1.000] & [0.507, 0.623, 0.742] & [0.000, 0.004, 0.344] \\
GSK3B & 450 & 150 & 150 & 150 & [0.000, 0.615, 1.000] & [0.760, 0.880, 1.000] & [0.510, 0.615, 0.750] & [0.000, 0.030, 0.380] \\
JNK3 & 450 & 150 & 150 & 150 & [0.000, 0.570, 1.000] & [0.760, 0.890, 1.000] & [0.510, 0.570, 0.750] & [0.000, 0.010, 0.340] \\
QED & 450 & 150 & 150 & 150 & [0.010, 0.644, 0.947] & [0.751, 0.819, 0.947] & [0.501, 0.644, 0.749] & [0.010, 0.348, 0.499] \\
\midrule
\rowcolor{lightgray} \multicolumn{9}{c}{\textbf{Target Based Design}} \\
Docking 5ht1b & 450 & 150 & 150 & 150 & [0.000, 0.607, 0.879] & [0.757, 0.771, 0.879] & [0.507, 0.607, 0.729] & [0.000, 0.450, 0.500] \\
Docking braf & 450 & 150 & 150 & 150 & [0.000, 0.600, 0.871] & [0.757, 0.771, 0.871] & [0.507, 0.600, 0.736] & [0.000, 0.464, 0.500] \\
Docking fa7 & 305 & 5 & 150 & 150 & [0.000, 0.507, 1.000] & [0.764, 0.800, 1.000] & [0.507, 0.536, 0.629] & [0.000, 0.450, 0.500] \\
Docking jak2 & 450 & 150 & 150 & 150 & [0.000, 0.586, 0.936] & [0.757, 0.771, 0.936] & [0.507, 0.586, 0.714] & [0.000, 0.464, 0.500] \\
Docking parp1 & 450 & 150 & 150 & 150 & [0.000, 0.614, 0.864] & [0.757, 0.771, 0.864] & [0.507, 0.614, 0.750] & [0.000, 0.471, 0.500] \\
\midrule
\rowcolor{lightgray} \multicolumn{9}{c}{\textbf{Material Design}} \\
Polymer CO$_2$ CH$_4$ & 160 & 1 & 9 & 150 & [0.042, 0.203, 0.777] & [0.777, 0.777, 0.777] & [0.538, 0.603, 0.731] & [0.042, 0.198, 0.487] \\
Polymer CO$_2$ N$_2$ & 158 & 0 & 8 & 150 & [0.000, 0.110, 0.631] &  & [0.506, 0.561, 0.631] & [0.000, 0.102, 0.498] \\
Polymer H$_2$ CH$_4$ & 177 & 13 & 14 & 150 & [0.000, 0.117, 1.000] & [0.752, 0.988, 1.000] & [0.509, 0.615, 0.739] & [0.000, 0.050, 0.487] \\
Polymer H$_2$ N$_2$ & 175 & 7 & 18 & 150 & [0.005, 0.178, 1.000] & [0.783, 0.974, 1.000] & [0.514, 0.619, 0.731] & [0.005, 0.151, 0.495] \\
Polymer O$_2$ N$_2$ & 173 & 0 & 23 & 150 & [0.309, 0.355, 0.726] &  & [0.501, 0.579, 0.726] & [0.309, 0.346, 0.493] \\
\bottomrule
\end{tabular}
\end{adjustbox}
\end{table}

We curate 33 benchmark tasks across six categories to evaluate \method against 13 baselines, including eight molecular optimization methods, two conditional generation models, and three LLMs. Details are in~\cref{tab:benchmark_task}.
The benchmarks span seven drug rediscovery tasks, seven drug multi-objective optimization (MPO) tasks, five structure-constrained generation tasks, four drug design tasks, five target-based generation tasks, and five polymer property design tasks. Specifically, the benchmarks span the following tasks:
\begin{itemize}
    \item \textbf{Drug rediscovery (7 tasks):} Celecoxib rediscovery, Mestranol similarity, Thiothixene rediscovery, Troglitazone rediscovery, Median 1 (median similarity between camphor and menthol), Median 2 (median similarity between tadalafil and sildenafil), and Albuterol similarity. These tasks use Oracle scoring functions based on the similarity between the drug and the target using extended connectivity fingerprint~\citep{brown2019guacamol}.
    
    \item \textbf{Drug MPO (7 tasks):} Perindopril MPO, Ranolazine MPO, Osimertinib MPO, Zaleplon MPO, Sitagliptin MPO, Amlodipine MPO, and Fexofenadine MPO.
    These tasks use Oracle scoring functions based on drug–target similarity with extended connectivity fingerprints, along with additional constraints such as logP, TPSA, and Bertz, computed using RDKit~\citep{brown2019guacamol}.
    
    \item \textbf{Structure-constrained design (5 tasks):} Isomers \ce{C7H8N2O2}, Isomers \ce{C9H10N2O2PF2Cl}, Decoration hop, Scaffold hop, and Valsartan SMARTS. These tasks use Oracle scoring functions primarily based on SMARTS patterns that evaluate whether a particular structure is present or absent in the target, optionally combined with other computational constraints. Or whether the target is an isomer of the molecular formula.
    
    \item \textbf{Drug design (4 tasks):} DRD2, JNK3, GSK3$\beta$, and QED. These tasks use Oracle scoring functions based on ML models. QED is based on RDKit.
    
    \item \textbf{Target-based design (5 tasks):} Docking BRAF, Docking PARP1, Docking JAK2, Docking FA7, and Docking 5-HT1B. These tasks use Oracle scoring functions based on the docking program QuickVina 2~\citep{alhossary2015fast}. Docking scores are negative, with smaller values indicating better binding. To map them into $[0,1]$ (where larger values are better), we use $s = \operatorname{clip}\!\left(-\tfrac{\text{docking score}}{14}, 0, 1\right)$, where $\operatorname{clip}(x,a,b) = \min(\max(x,a),b)$.
    
    \item \textbf{Material design (5 tasks):} Polymer gas separation for different gas pairs: CO\textsubscript{2}/CH\textsubscript{4}, CO\textsubscript{2}/N\textsubscript{2}, H\textsubscript{2}/CH\textsubscript{4}, H\textsubscript{2}/N\textsubscript{2}, and O\textsubscript{2}/N\textsubscript{2}. Each task studies whether two gases can be separated based on the polymeric membrane materials. We evaluate their selectivity score for gas separation~\citep{robeson2008upper}, defined as the log-ratio of permeabilities relative to an empirical boundary, shifted and clipped into the range $[0,1]$. Gas permeabilities are calculated using ML models trained on all available labeled data (a superset of the task-specific data), and the selectivity score is then computed based on gas permeabilities.
\end{itemize}

Each task contains up to 450 molecule–score pairs, evenly split into positive, medium, and negative groups. Some tasks may have fewer pairs due to insufficient positive examples, as shown in~\cref{tab:benchmark_task}. For instance, the Median 1 and Median 2 tasks in drug rediscovery have no positive or medium examples.
These pairs are used to train predictors for molecular optimization methods, conditional generators, or to provide demonstrations for ICL methods. Task scores lie within $[0,1]$, with the objective of generating molecules with score $1$. Each task also defines an oracle function, which is used only for evaluation, except by molecular optimization methods that actively query the oracle.

For baselines, we compare against four molecular optimization methods from the PMO benchmark~\citep{gao2022sample}. They are the top four methods selected from 25 candidates: Graph Genetic Algorithm (GraphGA), REINVENT (SMILES-based), Gaussian Process Bayesian Optimization (GPBO), and Superfast Traversal, Optimization, Novelty, Exploration, and Discovery (STONED, based on SELFIES). SELFIES is unavailable for polymers and STONED cannot be applied to material design tasks. We have two evaluation settings: one with 100 oracle calls and one with 10,000 predictor calls. While PMO permits up to 10,000 Oracle calls, such budgets are impractical in real-world settings due to the cost and time associated with laboratory experiments, which may require days to months for a single call. To address this issue, we examine whether molecular optimization methods can be paired with predictor calls. Following prior work~\citep{gao2022sample,liu2024graph}, we use a random forest predictor trained on all 450 molecule–score pairs as the task-specific predictor.

We include conditional generation models such as LSTM and Graph DiT~\citep{liu2024graph}. They are trained on all available training data for each task. For ICL, we compare \method (739M parameters) with recent large-scale LLMs, including DeepSeek-V3~\citep{liu2024deepseek}, GPT-4o~\citep{achiam2023gpt}, and Qwen-Max~\citep{yang2025qwen2}, each with up to hundreds of billions of parameters.
For LLMs, we sample 12 positive, 6 medium, and 6 negative as demonstrations.

For \method, we set the context size to 150 motif tokens. Excluding the target molecule, the context includes on average 23 demonstrations: half positive, one quarter medium, and one quarter negative. For evaluation, each method generates 100 valid, unique, and novel molecules per task, which are scored by oracle functions. We report the average of the top-10 oracle scores as the performance score and compute its harmonic mean with the diversity score. The diversity score is computed as 
\begin{equation}\label{eq:int-div}
\text{IntDiv}(G) \;=\; 1 \;-\; 
\left( \frac{1}{|G|^{2}} \sum_{\substack{m_{1}, m_{2} \in G \\ m_{1} \neq m_{2}}} 
T(m_{1}, m_{2})^2 \right)^{\tfrac{1}{2}},
\end{equation}
where $G$ denotes the generated set of molecules for evaluation.
For \method, we first generate 1000 candidates and select the top 100 with the highest consistency scores, prioritizing alignment with the context order of positive, medium, and negative examples.

\begin{table}[t]
\centering
\caption{Top-1 performance across 33 tasks. Scores are reported with a target of 1 as mean ± std by category. Best results in each column is \textbf{bolded}.
}
\label{tab:overall_top1}
\resizebox{\textwidth}{!}{%
\begin{tabular}{lcccccccc}
\toprule
Task Category & Drug & Drug & Structure & Drug & Target & Material & Avg & Total \\
 & Rediscovery & MPO & Constrained & Design & Based & Design & Rank & Sum \\
\# Tasks & 7 & 7 & 5 & 4 & 5 & 5 & 33 & 33 \\
\midrule
\multicolumn{9}{c}{Molecular Optimization Methods with 100 Oracle Calls} \\
\midrule
GraphGA & 0.28±0.06 & 0.49±0.18 & 0.46±0.14 & 0.45±0.33 & 0.74±0.04 & 0.72±0.23 & 6.65 & 16.73 \\
REINVENT & 0.31±0.07 & 0.47±0.16 & 0.45±0.14 & 0.56±0.38 & 0.75±0.06 & 0.00±0.00 & 7.55 & 13.68 \\
GPBO & 0.28±0.06 & 0.46±0.18 & 0.47±0.14 & 0.41±0.35 & 0.75±0.07 & 0.80±0.25 & 7.03 & 16.95 \\
STONED & 0.28±0.06 & 0.49±0.18 & 0.46±0.14 & 0.44±0.35 & 0.74±0.06 & \scriptsize NO SELFIES & 8.31 & 13.07 \\
\midrule
\multicolumn{9}{c}{Molecular Optimization Methods with 10000 Predictor Calls} \\
\midrule
GraphGA & 0.33±0.14 & 0.56±0.21 & 0.54±0.38 & 0.60±0.41 & 0.83±0.12 & 0.70±0.21 & 6.04 & 19.03 \\
REINVENT & 0.36±0.29 & 0.38±0.30 & 0.42±0.43 & 0.77±0.12 & 0.59±0.37 & 0.87±0.17 & 6.73 & 17.67 \\
GPBO & 0.37±0.31 & 0.50±0.25 & 0.47±0.32 & 0.64±0.38 & \textbf{0.84±0.08} & 0.58±0.34 & 7.06 & 18.13 \\
STONED & 0.28±0.17 & 0.43±0.25 & 0.51±0.32 & 0.30±0.10 & 0.26±0.36 & \scriptsize NO SELFIES & 9.93 & 10.07 \\
\midrule
\multicolumn{9}{c}{Conditional Generation Models} \\
\midrule
LSTM & 0.47±0.36 & 0.33±0.20 & \textbf{0.64±0.39} & 0.36±0.34 & 0.73±0.12 & 0.48±0.29 & 8.76 & 16.30 \\
Graph-DiT & 0.46±0.27 & 0.53±0.07 & 0.60±0.36 & 0.51±0.44 & 0.70±0.07 & 0.78±0.23 & 7.61 & 19.44 \\
\midrule
\multicolumn{9}{c}{Learning from In-Context Demonstrations} \\
\midrule
DeepSeek-V3 & \textbf{0.66±0.37} & \textbf{0.60±0.26} & 0.54±0.25 & 0.74±0.14 & 0.71±0.12 & 0.75±0.27 & 6.34 & 21.77 \\
GPT-4o & 0.53±0.31 & 0.56±0.20 & 0.52±0.32 & 0.54±0.41 & 0.69±0.09 & 0.77±0.22 & 7.34 & 19.69 \\
Qwen-Max & 0.18±0.28 & 0.19±0.17 & 0.53±0.34 & 0.51±0.45 & 0.21±0.29 & 0.25±0.40 & 10.25 & 9.60 \\
DemoDiff (Ours) & 0.54±0.33 & 0.54±0.19 & 0.59±0.37 & \textbf{0.91±0.07} & 0.77±0.10 & \textbf{0.93±0.16} & \textbf{3.94} & \textbf{22.63} \\
\bottomrule
\end{tabular}
}
\end{table}

\subsection{Additional Discussion of Experimental Results}\label{addsec:disc-result}

We include more results in~\cref{tab:drug_design_top10_harmonic,tab:drug_design_top1,tab:drug_mpo_top10_harmonic,tab:drug_mpo_top1,tab:material_design_top10_harmonic,tab:material_design_top1,tab:material_design_top10_harmonic,tab:structure_constrained_design_top1,tab:structure_constrained_design_top10_harmonic}. Beyond the discussion of ICL methods and \method in~\cref{sec:perf-discussion}, we have additional observations:

\textbf{Oracle quality critically affects molecular optimization.} Comparing molecular optimization methods under varying numbers of function calls, we find that allowing more predictor queries does not consistently lead to better performance. This suggests that both the quantity and quality of function evaluations (oracle or predictor) are essential for guiding molecular optimization. While not the main focus of this study, this insight points to an important direction for future work. For instance, in the structure-constrained design task involving the Valsartan SMARTS pattern (\texttt{CN(C=O)Cc1ccc(c2ccccc2)cc1}), shown in~\cref{tab:structure_constrained_design_top1}, all molecular optimization methods receive a score of zero. This failure is due to a predictor trained on limited data, which cannot model the latent design constraints, such as satisfying multiple SMARTS patterns and physicochemical properties (e.g., logP, TPSA, and Bertz index~\citep{brown2019guacamol}). In contrast, in target-based design tasks (e.g.,~\cref{tab:target_based_design_top1}), where training data are sufficient, more predictor calls improve performance by allowing finer structural optimization.

\textbf{Performance alignment across methods may indicate task difficulty.} It is challenging to formally quantify task difficulty, as it depends on the oracle definition and data quality. However, we observe that tasks where molecular optimization performs well—such as target-based or material design—are also more tractable for ICL methods, which can infer the underlying concept with few demonstrations. This alignment suggests that task difficulty may be partially reflected in cross-method consistency. Nonetheless, exceptions exist. As shown in~\cref{tab:drug_design_top1}, for DRD2 and JNK3, molecular optimization underperforms under limited supervision, while ICL methods, including DeepSeek-V3, GPT-4o, and \method, achieve strong results.

\begin{table}[t]
\centering
\caption{Harmonic mean of Top-10 performance and diversity scores on the Drug Design task category. Scores are reported with a target of 1. Best results in each column is \textbf{bolded}.}
\label{tab:drug_design_top10_harmonic}
\small
\begin{tabular}{lcccccc}
\toprule
Task & DRD2 & JNK3 & GSK3B & QED & Avg & Total \\
 &  &  &  &  & Rank & Sum \\
\midrule
\multicolumn{7}{c}{Molecular Optimization Methods with 100 Oracle Calls} \\
\midrule
GraphGA & 0.22 & 0.23 & 0.31 & 0.90 & 6.25 & 1.65 \\
REINVENT & 0.31 & 0.21 & 0.26 & \textbf{0.90} & 7.00 & 1.68 \\
GPBO & 0.20 & 0.25 & 0.24 & 0.89 & 7.75 & 1.57 \\
STONED & 0.22 & 0.23 & 0.31 & 0.90 & 7.25 & 1.65 \\
\midrule
\multicolumn{7}{c}{Molecular Optimization Methods with 10000 Predictor Calls} \\
\midrule
GraphGA & 0.60 & 0.18 & 0.40 & 0.79 & 7.00 & 1.97 \\
REINVENT & 0.31 & 0.09 & 0.40 & 0.71 & 8.75 & 1.51 \\
GPBO & 0.55 & 0.27 & 0.54 & 0.60 & 6.00 & 1.97 \\
STONED & 0.14 & 0.23 & 0.37 & 0.32 & 9.50 & 1.07 \\
\midrule
\multicolumn{7}{c}{Conditional Generation Models} \\
\midrule
LSTM & 0.15 & 0.04 & 0.27 & 0.83 & 11.00 & 1.30 \\
Graph-DiT & 0.78 & 0.08 & 0.23 & 0.81 & 9.50 & 1.90 \\
\midrule
\multicolumn{7}{c}{Learning from In-Context Demonstrations} \\
\midrule
DeepSeek-V3 & 0.71 & \textbf{0.65} & 0.42 & 0.84 & 3.75 & 2.62 \\
GPT-4o & 0.80 & 0.13 & 0.13 & 0.84 & 8.25 & 1.90 \\
Qwen-Max & 0.00 & 0.14 & 0.36 & 0.68 & 10.75 & 1.18 \\
DemoDiff (Ours) & \textbf{0.88} & \textbf{0.65} & \textbf{0.78} & 0.87 & \textbf{2.25} & \textbf{3.18} \\
\bottomrule
\end{tabular}
\end{table}
\begin{table}[t]
\centering
\caption{Top-1 performance on the Drug Design task category. Scores are reported with a target of 1. Best results in each column is \textbf{bolded}.}
\label{tab:drug_design_top1}
\small
\begin{tabular}{lcccccc}
\toprule
Task & DRD2 & JNK3 & GSK3B & QED & Avg & Total \\
 &  &  &  &  & Rank & Sum \\
\midrule
\multicolumn{7}{c}{Molecular Optimization Methods with 100 Oracle Calls} \\
\midrule
GraphGA & 0.23 & 0.23 & 0.41 & 0.94 & 6.50 & 1.82 \\
REINVENT & 0.83 & 0.23 & 0.24 & \textbf{0.94} & 6.00 & 2.25 \\
GPBO & 0.23 & 0.23 & 0.23 & 0.94 & 8.25 & 1.64 \\
STONED & 0.19 & 0.23 & 0.41 & 0.94 & 8.25 & 1.77 \\
\midrule
\multicolumn{7}{c}{Molecular Optimization Methods with 10000 Predictor Calls} \\
\midrule
GraphGA & 0.99 & 0.14 & 0.38 & 0.90 & 8.00 & 2.41 \\
REINVENT & 0.76 & 0.61 & 0.80 & 0.91 & 5.25 & 3.08 \\
GPBO & 0.99 & 0.19 & 0.47 & 0.93 & 6.50 & 2.58 \\
STONED & 0.25 & 0.21 & 0.44 & 0.30 & 9.75 & 1.20 \\
\midrule
\multicolumn{7}{c}{Conditional Generation Models} \\
\midrule
LSTM & 0.28 & 0.06 & 0.26 & 0.85 & 11.00 & 1.45 \\
Graph-DiT & 0.99 & 0.07 & 0.22 & 0.78 & 10.25 & 2.06 \\
\midrule
\multicolumn{7}{c}{Learning from In-Context Demonstrations} \\
\midrule
DeepSeek-V3 & 0.77 & 0.75 & 0.54 & 0.88 & 5.75 & 2.94 \\
GPT-4o & 0.94 & 0.21 & 0.17 & 0.85 & 10.25 & 2.17 \\
Qwen-Max & 0.00 & 0.26 & \textbf{0.89} & 0.89 & 7.25 & 2.04 \\
DemoDiff (Ours) & \textbf{1.00} & \textbf{0.83} & \textbf{0.89} & 0.93 & \textbf{2.00} & \textbf{3.65} \\
\bottomrule
\end{tabular}
\end{table}
\begin{table}[t]
\centering
\caption{Harmonic mean of Top-10 performance and diversity scores on the Drug MPO task category. Scores are reported with a target of 1. Best results in each column is \textbf{bolded}.}
\label{tab:drug_mpo_top10_harmonic}
\resizebox{\textwidth}{!}{%
\begin{tabular}{lccccccccc}
\toprule
Task & Perindopril & Ranolazine & Osimertinib & Zaleplon & Sitagliptin & Amlodipine & Fexofenadine & Avg & Total \\
 & MPO & MPO & MPO & MPO & MPO & MPO & MPO & Rank & Sum \\
\midrule
\multicolumn{10}{c}{Molecular Optimization Methods with 100 Oracle Calls} \\
\midrule
GraphGA & 0.52 & 0.35 & 0.77 & 0.47 & 0.24 & 0.60 & 0.70 & \textbf{4.29} & 3.64 \\
REINVENT & 0.50 & 0.39 & 0.75 & \textbf{0.48} & 0.24 & 0.59 & 0.66 & 5.14 & 3.62 \\
GPBO & 0.51 & 0.39 & 0.76 & 0.45 & 0.23 & 0.59 & 0.68 & 6.29 & 3.60 \\
STONED & 0.52 & 0.35 & 0.77 & 0.47 & 0.24 & 0.60 & 0.70 & 5.29 & 3.64 \\
\midrule
\multicolumn{10}{c}{Molecular Optimization Methods with 10000 Predictor Calls} \\
\midrule
GraphGA & 0.41 & 0.60 & 0.64 & 0.48 & 0.13 & 0.59 & 0.62 & 7.00 & 3.48 \\
REINVENT & 0.11 & 0.24 & 0.65 & 0.39 & 0.00 & 0.09 & 0.10 & 12.57 & 1.58 \\
GPBO & 0.23 & 0.36 & 0.68 & 0.45 & 0.12 & 0.62 & 0.67 & 7.43 & 3.12 \\
STONED & 0.53 & 0.26 & 0.47 & 0.47 & 0.01 & 0.60 & 0.45 & 8.00 & 2.79 \\
\midrule
\multicolumn{10}{c}{Conditional Generation Models} \\
\midrule
LSTM & 0.06 & 0.14 & 0.26 & 0.09 & 0.13 & 0.21 & 0.21 & 12.57 & 1.10 \\
Graph-DiT & 0.59 & 0.24 & 0.71 & 0.43 & \textbf{0.31} & 0.55 & 0.66 & 7.00 & 3.49 \\
\midrule
\multicolumn{10}{c}{Learning from In-Context Demonstrations} \\
\midrule
DeepSeek-V3 & 0.49 & \textbf{0.63} & 0.70 & 0.42 & 0.10 & 0.63 & 0.58 & 7.14 & 3.55 \\
GPT-4o & \textbf{0.63} & 0.39 & 0.68 & 0.46 & 0.17 & \textbf{0.69} & 0.67 & 4.71 & 3.70 \\
Qwen-Max & 0.26 & 0.21 & 0.00 & 0.39 & 0.00 & 0.25 & 0.06 & 12.86 & 1.17 \\
DemoDiff (Ours) & 0.52 & 0.58 & \textbf{0.80} & 0.43 & 0.09 & 0.62 & \textbf{0.73} & 4.71 & \textbf{3.78} \\
\bottomrule
\end{tabular}
}
\end{table}
\begin{table}[t]
\centering
\caption{Top-1 performance on the Drug MPO task category. Scores are reported with a target of 1. Best results in each column is \textbf{bolded}.}
\label{tab:drug_mpo_top1}
\resizebox{\textwidth}{!}{%
\begin{tabular}{lccccccccc}
\toprule
Task & Perindopril & Ranolazine & Osimertinib & Zaleplon & Sitagliptin & Amlodipine & Fexofenadine & Avg & Total \\
 & MPO & MPO & MPO & MPO & MPO & MPO & MPO & Rank & Sum \\
\midrule
\multicolumn{10}{c}{Molecular Optimization Methods with 100 Oracle Calls} \\
\midrule
GraphGA & 0.43 & 0.36 & 0.78 & 0.40 & 0.23 & 0.56 & 0.62 & 5.86 & 3.40 \\
REINVENT & 0.40 & 0.41 & 0.74 & 0.43 & 0.23 & 0.48 & 0.58 & 8.14 & 3.26 \\
GPBO & 0.38 & 0.36 & 0.78 & 0.37 & 0.23 & 0.48 & 0.60 & 9.71 & 3.20 \\
STONED & 0.43 & 0.36 & 0.78 & 0.40 & 0.23 & 0.56 & 0.62 & 7.29 & 3.40 \\
\midrule
\multicolumn{10}{c}{Molecular Optimization Methods with 10000 Predictor Calls} \\
\midrule
GraphGA & 0.40 & 0.67 & \textbf{0.83} & 0.41 & 0.28 & 0.53 & \textbf{0.81} & \textbf{4.14} & 3.93 \\
REINVENT & 0.17 & 0.34 & 0.82 & 0.40 & 0.00 & 0.22 & 0.72 & 9.43 & 2.68 \\
GPBO & 0.15 & 0.41 & 0.78 & 0.40 & 0.30 & 0.64 & 0.80 & 6.00 & 3.48 \\
STONED & 0.44 & 0.26 & 0.77 & 0.42 & 0.01 & 0.52 & 0.62 & 8.57 & 3.03 \\
\midrule
\multicolumn{10}{c}{Conditional Generation Models} \\
\midrule
LSTM & 0.08 & 0.11 & 0.59 & 0.22 & 0.44 & 0.42 & 0.46 & 11.71 & 2.31 \\
Graph-DiT & 0.58 & 0.51 & 0.61 & 0.46 & \textbf{0.46} & 0.48 & 0.61 & 6.00 & 3.71 \\
\midrule
\multicolumn{10}{c}{Learning from In-Context Demonstrations} \\
\midrule
DeepSeek-V3 & \textbf{0.74} & \textbf{0.77} & 0.78 & 0.39 & 0.10 & \textbf{0.77} & 0.63 & 5.57 & \textbf{4.17} \\
GPT-4o & 0.74 & 0.47 & 0.71 & 0.43 & 0.19 & 0.71 & 0.65 & 5.57 & 3.90 \\
Qwen-Max & 0.30 & 0.21 & 0.00 & \textbf{0.49} & 0.00 & 0.22 & 0.12 & 11.29 & 1.33 \\
DemoDiff (Ours) & 0.44 & 0.65 & 0.77 & 0.39 & 0.24 & 0.55 & 0.73 & 5.71 & 3.77 \\
\bottomrule
\end{tabular}
}
\end{table}
\begin{table}[t]
\centering
\caption{Harmonic mean of Top-10 performance and diversity scores on the Drug Rediscovery task category. Scores are reported with a target of 1. Best results in each column is \textbf{bolded}.}
\label{tab:drug_rediscovery_top10_harmonic}
\resizebox{\textwidth}{!}{%
\begin{tabular}{lccccccccc}
\toprule
Task & Celecoxib & Mestranol & Thiothixene & Troglitazone & Median 1 & Median 2 & Albuterol & Avg & Total \\
 & Rediscovery & Similarity & Rediscovery & Rediscovery &  &  & Similarity & Rank & Sum \\
\midrule
\multicolumn{10}{c}{Molecular Optimization Methods with 100 Oracle Calls} \\
\midrule
GraphGA & 0.39 & 0.39 & 0.38 & 0.34 & 0.26 & 0.29 & 0.49 & 7.14 & 2.54 \\
REINVENT & 0.39 & 0.44 & 0.37 & 0.34 & 0.30 & 0.28 & 0.50 & 6.29 & 2.62 \\
GPBO & 0.37 & 0.39 & 0.39 & 0.33 & 0.28 & \textbf{0.30} & 0.51 & 6.57 & 2.57 \\
STONED & 0.39 & 0.39 & 0.38 & 0.34 & 0.26 & 0.29 & 0.49 & 8.14 & 2.54 \\
\midrule
\multicolumn{10}{c}{Molecular Optimization Methods with 10000 Predictor Calls} \\
\midrule
GraphGA & 0.38 & 0.41 & 0.38 & 0.35 & 0.28 & 0.28 & 0.55 & 6.86 & 2.61 \\
REINVENT & 0.31 & 0.15 & 0.38 & 0.26 & 0.19 & 0.27 & 0.53 & 9.29 & 2.08 \\
GPBO & 0.34 & 0.43 & 0.37 & 0.26 & 0.19 & 0.27 & 0.46 & 10.14 & 2.32 \\
STONED & 0.39 & 0.49 & 0.38 & 0.32 & 0.19 & 0.27 & 0.26 & 8.71 & 2.30 \\
\midrule
\multicolumn{10}{c}{Conditional Generation Models} \\
\midrule
LSTM & \textbf{0.61} & \textbf{0.66} & 0.20 & 0.25 & 0.20 & 0.12 & 0.70 & 7.71 & 2.74 \\
Graph-DiT & 0.60 & 0.58 & 0.33 & 0.22 & \textbf{0.39} & 0.17 & 0.73 & 6.57 & 3.02 \\
\midrule
\multicolumn{10}{c}{Learning from In-Context Demonstrations} \\
\midrule
DeepSeek-V3 & 0.54 & 0.46 & 0.47 & \textbf{0.57} & 0.18 & 0.25 & 0.70 & 5.29 & 3.18 \\
GPT-4o & 0.54 & 0.64 & \textbf{0.61} & 0.37 & 0.20 & 0.19 & 0.73 & \textbf{4.29} & \textbf{3.28} \\
Qwen-Max & 0.48 & 0.00 & 0.00 & 0.00 & 0.00 & 0.15 & 0.40 & 12.57 & 1.03 \\
DemoDiff (Ours) & 0.50 & 0.57 & 0.42 & 0.49 & 0.12 & 0.23 & \textbf{0.75} & 5.43 & 3.09 \\
\bottomrule
\end{tabular}
}
\end{table}
\begin{table}[t]
\centering
\caption{Top-1 performance on the Drug Rediscovery task category. Scores are reported with a target of 1. Best results in each column is \textbf{bolded}.}
\label{tab:drug_rediscovery_top1}
\resizebox{\textwidth}{!}{%
\begin{tabular}{lccccccccc}
\toprule
Task & Celecoxib & Mestranol & Thiothixene & Troglitazone & Median 1 & Median 2 & Albuterol & Avg & Total \\
 & Rediscovery & Similarity & Rediscovery & Rediscovery &  &  & Similarity & Rank & Sum \\
\midrule
\multicolumn{10}{c}{Molecular Optimization Methods with 100 Oracle Calls} \\
\midrule
GraphGA & 0.27 & 0.28 & 0.28 & 0.23 & 0.23 & \textbf{0.23} & 0.40 & 7.71 & 1.93 \\
REINVENT & 0.34 & 0.39 & 0.32 & 0.25 & 0.23 & \textbf{0.23} & 0.41 & 6.14 & 2.18 \\
GPBO & 0.28 & 0.31 & 0.29 & 0.23 & 0.23 & \textbf{0.23} & 0.41 & 7.43 & 1.99 \\
STONED & 0.27 & 0.28 & 0.28 & 0.23 & 0.23 & \textbf{0.23} & 0.40 & 9.43 & 1.93 \\
\midrule
\multicolumn{10}{c}{Molecular Optimization Methods with 10000 Predictor Calls} \\
\midrule
GraphGA & 0.35 & 0.38 & 0.28 & 0.25 & 0.21 & 0.20 & 0.62 & 7.29 & 2.30 \\
REINVENT & 0.36 & 0.28 & 0.28 & 0.27 & 0.13 & 0.20 & \textbf{1.00} & 7.29 & 2.51 \\
GPBO & 0.25 & 0.57 & 0.26 & 0.20 & 0.13 & 0.20 & \textbf{1.00} & 9.29 & 2.61 \\
STONED & 0.27 & 0.64 & 0.28 & 0.23 & 0.13 & 0.20 & 0.22 & 10.14 & 1.96 \\
\midrule
\multicolumn{10}{c}{Conditional Generation Models} \\
\midrule
LSTM & \textbf{1.00} & 0.82 & 0.18 & 0.33 & 0.20 & 0.09 & 0.69 & 6.86 & 3.31 \\
Graph-DiT & 0.78 & 0.47 & 0.51 & 0.18 & \textbf{0.40} & 0.10 & 0.81 & 6.43 & 3.25 \\
\midrule
\multicolumn{10}{c}{Learning from In-Context Demonstrations} \\
\midrule
DeepSeek-V3 & 0.77 & \textbf{1.00} & \textbf{0.84} & \textbf{0.75} & 0.12 & 0.17 & \textbf{1.00} & \textbf{4.71} & \textbf{4.64} \\
GPT-4o & 0.53 & 0.68 & 0.81 & 0.46 & 0.15 & 0.15 & 0.93 & 5.43 & 3.70 \\
Qwen-Max & 0.72 & 0.00 & 0.00 & 0.00 & 0.00 & 0.10 & 0.41 & 12.14 & 1.23 \\
DemoDiff (Ours) & 0.84 & 0.64 & 0.34 & 0.65 & 0.14 & 0.16 & \textbf{1.00} & \textbf{4.71} & 3.77 \\
\bottomrule
\end{tabular}
}
\end{table}
\begin{table}[t]
\centering
\caption{Harmonic mean of Top-10 performance and diversity scores on the Material Design task category. Scores are reported with a target of 1. Best results in each column is \textbf{bolded}.}
\label{tab:material_design_top10_harmonic}
\small
\begin{tabular}{lccccccc}
\toprule
Task & Polymer & Polymer & Polymer & Polymer & Polymer & Avg & Total \\
 & CO2/CH4 & CO2/N2 & H2/CH4 & H2/N2 & O2/N2 & Rank & Sum \\
\midrule
\multicolumn{8}{c}{Molecular Optimization Methods with 100 Oracle Calls} \\
\midrule
GraphGA & 0.59 & 0.42 & 0.71 & 0.52 & 0.64 & 4.60 & 2.88 \\
REINVENT & 0.00 & 0.00 & 0.00 & 0.00 & 0.00 & 11.00 & 0.00 \\
GPBO & 0.46 & 0.31 & \textbf{0.76} & 0.79 & 0.68 & 3.40 & 3.00 \\
STONED & \scriptsize NO SELFIES & \scriptsize NO SELFIES & \scriptsize NO SELFIES & \scriptsize NO SELFIES & \scriptsize NO SELFIES & \scriptsize NO SELFIES & \scriptsize NO SELFIES \\
\midrule
\multicolumn{8}{c}{Molecular Optimization Methods with 10000 Predictor Calls} \\
\midrule
GraphGA & 0.60 & \textbf{0.55} & 0.75 & 0.40 & 0.44 & 4.20 & 2.75 \\
REINVENT & 0.32 & 0.37 & 0.56 & 0.69 & 0.61 & 5.80 & 2.55 \\
GPBO & 0.55 & 0.00 & 0.57 & 0.52 & 0.46 & 7.20 & 2.09 \\
STONED & \scriptsize NO SELFIES & \scriptsize NO SELFIES & \scriptsize NO SELFIES & \scriptsize NO SELFIES & \scriptsize NO SELFIES & \scriptsize NO SELFIES & \scriptsize NO SELFIES \\
\midrule
\multicolumn{8}{c}{Conditional Generation Models} \\
\midrule
LSTM & 0.00 & 0.29 & 0.16 & 0.12 & 0.23 & 10.00 & 0.80 \\
Graph-DiT & 0.27 & 0.55 & 0.74 & 0.57 & 0.63 & 4.60 & 2.76 \\
\midrule
\multicolumn{8}{c}{Learning from In-Context Demonstrations} \\
\midrule
DeepSeek-V3 & 0.13 & 0.45 & 0.53 & 0.68 & 0.15 & 7.40 & 1.94 \\
GPT-4o & 0.37 & 0.30 & 0.44 & 0.70 & 0.33 & 7.00 & 2.14 \\
Qwen-Max & 0.00 & 0.09 & 0.00 & 0.00 & 0.42 & 10.80 & 0.51 \\
DemoDiff (Ours) & \textbf{0.63} & 0.52 & 0.72 & \textbf{0.81} & \textbf{0.68} & \textbf{2.00} & \textbf{3.36} \\
\bottomrule
\end{tabular}
\end{table}
\begin{table}[t]
\centering
\caption{Top-1 performance on the Material Design task category. Scores are reported with a target of 1. Best results in each column is \textbf{bolded}.}
\label{tab:material_design_top1}
\small
\begin{tabular}{lccccccc}
\toprule
Task & Polymer & Polymer & Polymer & Polymer & Polymer & Avg & Total \\
 & CO2/CH4 & CO2/N2 & H2/CH4 & H2/N2 & O2/N2 & Rank & Sum \\
\midrule
\multicolumn{8}{c}{Molecular Optimization Methods with 100 Oracle Calls} \\
\midrule
GraphGA & 0.79 & 0.41 & \textbf{1.00} & 0.57 & 0.84 & 5.60 & 3.60 \\
REINVENT & 0.00 & 0.00 & 0.00 & 0.00 & 0.00 & 11.20 & 0.00 \\
GPBO & 0.89 & 0.43 & \textbf{1.00} & \textbf{1.00} & 0.68 & 4.40 & 4.00 \\
STONED & \scriptsize NO SELFIES & \scriptsize NO SELFIES & \scriptsize NO SELFIES & \scriptsize NO SELFIES & \scriptsize NO SELFIES & \scriptsize NO SELFIES & \scriptsize NO SELFIES \\
\midrule
\multicolumn{8}{c}{Molecular Optimization Methods with 10000 Predictor Calls} \\
\midrule
GraphGA & 0.69 & 0.82 & \textbf{1.00} & 0.51 & 0.50 & 6.20 & 3.52 \\
REINVENT & 0.71 & 0.67 & \textbf{1.00} & \textbf{1.00} & 0.97 & 3.60 & 4.35 \\
GPBO & 0.67 & 0.00 & 0.80 & 0.85 & 0.57 & 8.40 & 2.90 \\
STONED & \scriptsize NO SELFIES & \scriptsize NO SELFIES & \scriptsize NO SELFIES & \scriptsize NO SELFIES & \scriptsize NO SELFIES & \scriptsize NO SELFIES & \scriptsize NO SELFIES \\
\midrule
\multicolumn{8}{c}{Conditional Generation Models} \\
\midrule
LSTM & 0.00 & 0.77 & 0.55 & 0.63 & 0.44 & 8.60 & 2.39 \\
Graph-DiT & 0.44 & \textbf{1.00} & \textbf{1.00} & 0.79 & 0.69 & 5.60 & 3.92 \\
\midrule
\multicolumn{8}{c}{Learning from In-Context Demonstrations} \\
\midrule
DeepSeek-V3 & 0.35 & 0.60 & 0.95 & \textbf{1.00} & 0.86 & 6.20 & 3.76 \\
GPT-4o & 0.67 & 0.52 & \textbf{1.00} & \textbf{1.00} & 0.65 & 6.80 & 3.85 \\
Qwen-Max & 0.00 & 0.34 & 0.00 & 0.02 & 0.91 & 9.60 & 1.27 \\
DemoDiff (Ours) & \textbf{1.00} & 0.64 & \textbf{1.00} & \textbf{1.00} & \textbf{1.00} & \textbf{1.80} & \textbf{4.64} \\
\bottomrule
\end{tabular}
\end{table}
\begin{table}[t]
\centering
\caption{Top-1 performance on the Structure Constrained Design task category. Scores are reported with a target of 1. Best results in each column is \textbf{bolded}.}
\label{tab:structure_constrained_design_top1}
\small
\begin{tabular}{lccccccc}
\toprule
Task & Isomers & Isomers & Deco Hop & Scaffold & Valsartan & Avg & Total \\
 & c7h8n2o2 & c9h10n2o2pf2cl &  & Hop & Smarts & Rank & Sum \\
\midrule
\multicolumn{8}{c}{Molecular Optimization Methods with 100 Oracle Calls} \\
\midrule
GraphGA & 0.55 & 0.47 & 0.58 & 0.45 & \textbf{0.23} & 7.60 & 2.29 \\
REINVENT & 0.55 & 0.44 & 0.58 & 0.45 & \textbf{0.23} & 8.20 & 2.26 \\
GPBO & 0.55 & 0.50 & 0.59 & 0.47 & \textbf{0.23} & 6.80 & 2.34 \\
STONED & 0.55 & 0.47 & 0.58 & 0.45 & \textbf{0.23} & 9.40 & 2.29 \\
\midrule
\multicolumn{8}{c}{Molecular Optimization Methods with 10000 Predictor Calls} \\
\midrule
GraphGA & \textbf{1.00} & 0.78 & 0.54 & 0.39 & 0.00 & 8.00 & 2.72 \\
REINVENT & \textbf{1.00} & 0.00 & 0.62 & 0.47 & 0.00 & 6.60 & 2.09 \\
GPBO & 0.88 & 0.47 & 0.55 & 0.45 & 0.00 & 10.20 & 2.36 \\
STONED & 0.82 & 0.72 & 0.57 & 0.47 & 0.00 & 8.40 & 2.57 \\
\midrule
\multicolumn{8}{c}{Conditional Generation Models} \\
\midrule
LSTM & \textbf{1.00} & \textbf{0.87} & 0.54 & \textbf{0.78} & 0.00 & 6.20 & \textbf{3.18} \\
Graph-DiT & 0.88 & 0.82 & 0.53 & 0.77 & 0.00 & 7.80 & 3.01 \\
\midrule
\multicolumn{8}{c}{Learning from In-Context Demonstrations} \\
\midrule
DeepSeek-V3 & 0.72 & 0.50 & 0.84 & 0.41 & 0.21 & 7.80 & 2.68 \\
GPT-4o & 0.74 & 0.82 & 0.59 & 0.46 & 0.00 & 6.20 & 2.61 \\
Qwen-Max & 0.90 & 0.73 & 0.56 & 0.45 & 0.01 & 7.40 & 2.67 \\
DemoDiff (Ours) & 0.88 & 0.73 & \textbf{0.86} & 0.49 & 0.00 & \textbf{4.40} & 2.96 \\
\bottomrule
\end{tabular}
\end{table}
\begin{table}[t]
\centering
\caption{Harmonic mean of Top-10 performance and diversity scores on the Structure Constrained Design task category. Scores are reported with a target of 1. Best results in each column is \textbf{bolded}.}
\label{tab:structure_constrained_design_top10_harmonic}
\small
\begin{tabular}{lccccccc}
\toprule
Task & Isomers & Isomers & Deco Hop & Scaffold & Valsartan & Avg & Total \\
 & c7h8n2o2 & c9h10n2o2pf2cl &  & Hop & Smarts & Rank & Sum \\
\midrule
\multicolumn{8}{c}{Molecular Optimization Methods with 100 Oracle Calls} \\
\midrule
GraphGA & 0.30 & 0.44 & 0.68 & 0.57 & \textbf{0.14} & 6.40 & 2.13 \\
REINVENT & 0.36 & 0.40 & 0.68 & 0.57 & 0.14 & 7.00 & 2.15 \\
GPBO & 0.27 & 0.45 & 0.68 & 0.57 & 0.14 & 6.00 & 2.11 \\
STONED & 0.30 & 0.44 & 0.68 & 0.57 & \textbf{0.14} & 7.40 & 2.13 \\
\midrule
\multicolumn{8}{c}{Molecular Optimization Methods with 10000 Predictor Calls} \\
\midrule
GraphGA & 0.71 & 0.69 & 0.42 & 0.45 & 0.00 & 8.60 & 2.27 \\
REINVENT & 0.49 & 0.00 & 0.46 & 0.30 & 0.00 & 12.00 & 1.24 \\
GPBO & 0.69 & 0.29 & 0.65 & 0.53 & 0.00 & 10.00 & 2.16 \\
STONED & 0.62 & 0.65 & 0.68 & 0.57 & 0.00 & 7.20 & 2.51 \\
\midrule
\multicolumn{8}{c}{Conditional Generation Models} \\
\midrule
LSTM & 0.80 & 0.75 & 0.65 & 0.55 & 0.00 & 7.20 & 2.75 \\
Graph-DiT & \textbf{0.85} & 0.74 & 0.66 & \textbf{0.66} & 0.00 & 5.60 & \textbf{2.91} \\
\midrule
\multicolumn{8}{c}{Learning from In-Context Demonstrations} \\
\midrule
DeepSeek-V3 & 0.64 & 0.47 & \textbf{0.70} & 0.52 & 0.09 & 6.40 & 2.43 \\
GPT-4o & 0.61 & \textbf{0.78} & 0.67 & 0.54 & 0.00 & 6.80 & 2.60 \\
Qwen-Max & 0.70 & 0.55 & 0.00 & 0.36 & 0.01 & 9.00 & 1.62 \\
DemoDiff (Ours) & 0.82 & 0.77 & 0.64 & 0.57 & 0.00 & \textbf{5.40} & 2.80 \\
\bottomrule
\end{tabular}
\end{table}
\begin{table}[t]
\centering
\caption{Top-1 performance on the Target Based Design task category. Scores are reported with a target of 1. Best results in each column is \textbf{bolded}.}
\label{tab:target_based_design_top1}
\small
\begin{tabular}{lccccccc}
\toprule
Task & Docking & Docking & Docking & Docking & Docking & Avg & Total \\
 & Braf & Parp1 & Jak2 & Fa7 & 5HT1B & Rank & Sum \\
\midrule
\multicolumn{8}{c}{Molecular Optimization Methods with 100 Oracle Calls} \\
\midrule
GraphGA & 0.75 & 0.76 & 0.74 & 0.67 & 0.76 & 6.60 & 3.69 \\
REINVENT & 0.77 & 0.80 & 0.69 & 0.66 & 0.80 & 5.60 & 3.73 \\
GPBO & 0.79 & 0.83 & 0.72 & 0.64 & 0.78 & 5.60 & 3.76 \\
STONED & 0.74 & 0.77 & 0.77 & 0.64 & 0.76 & 7.20 & 3.69 \\
\midrule
\multicolumn{8}{c}{Molecular Optimization Methods with 10000 Predictor Calls} \\
\midrule
GraphGA & \textbf{0.84} & 0.91 & 0.79 & 0.66 & \textbf{0.95} & 2.60 & 4.15 \\
REINVENT & 0.68 & \textbf{1.00} & 0.59 & 0.69 & 0.00 & 8.20 & 2.96 \\
GPBO & 0.84 & 0.91 & \textbf{0.84} & \textbf{0.71} & 0.91 & \textbf{2.00} & \textbf{4.21} \\
STONED & 0.71 & 0.00 & 0.61 & 0.00 & 0.00 & 12.80 & 1.32 \\
\midrule
\multicolumn{8}{c}{Conditional Generation Models} \\
\midrule
LSTM & 0.74 & 0.76 & 0.63 & 0.61 & 0.92 & 8.20 & 3.66 \\
Graph-DiT & 0.72 & 0.77 & 0.66 & 0.59 & 0.74 & 9.60 & 3.49 \\
\midrule
\multicolumn{8}{c}{Learning from In-Context Demonstrations} \\
\midrule
DeepSeek-V3 & 0.80 & 0.72 & 0.81 & 0.52 & 0.72 & 8.00 & 3.57 \\
GPT-4o & 0.77 & 0.72 & 0.64 & 0.57 & 0.76 & 9.80 & 3.46 \\
Qwen-Max & 0.54 & 0.00 & 0.00 & 0.51 & 0.00 & 13.80 & 1.06 \\
DemoDiff (Ours) & 0.81 & 0.86 & 0.79 & 0.60 & 0.77 & 5.00 & 3.84 \\
\bottomrule
\end{tabular}
\end{table}
\begin{table}[t]
\centering
\caption{Harmonic mean of Top-10 performance and diversity scores on the Target Based Design task category. Scores are reported with a target of 1. Best results in each column is \textbf{bolded}.}
\label{tab:target_based_design_top10_harmonic}
\small
\begin{tabular}{lccccccc}
\toprule
Task & Docking & Docking & Docking & Docking & Docking & Avg & Total \\
 & Braf & Parp1 & Jak2 & Fa7 & 5HT1B & Rank & Sum \\
\midrule
\multicolumn{8}{c}{Molecular Optimization Methods with 100 Oracle Calls} \\
\midrule
GraphGA & 0.78 & 0.79 & 0.76 & 0.70 & 0.79 & 2.80 & 3.82 \\
REINVENT & 0.77 & 0.79 & 0.73 & 0.71 & 0.78 & 3.80 & 3.78 \\
GPBO & 0.77 & 0.79 & 0.76 & 0.70 & 0.78 & 3.80 & 3.79 \\
STONED & 0.77 & 0.78 & 0.77 & 0.70 & 0.78 & 4.00 & 3.80 \\
\midrule
\multicolumn{8}{c}{Molecular Optimization Methods with 10000 Predictor Calls} \\
\midrule
GraphGA & 0.66 & 0.70 & 0.60 & 0.57 & 0.70 & 10.40 & 3.22 \\
REINVENT & 0.19 & 0.27 & 0.31 & 0.09 & 0.00 & 12.80 & 0.86 \\
GPBO & 0.69 & 0.78 & 0.76 & \textbf{0.73} & 0.74 & 6.00 & 3.70 \\
STONED & 0.48 & 0.00 & 0.52 & 0.00 & 0.00 & 12.80 & 1.00 \\
\midrule
\multicolumn{8}{c}{Conditional Generation Models} \\
\midrule
LSTM & 0.73 & 0.73 & 0.71 & 0.66 & 0.77 & 7.40 & 3.61 \\
Graph-DiT & 0.71 & 0.74 & 0.70 & 0.66 & 0.75 & 8.20 & 3.55 \\
\midrule
\multicolumn{8}{c}{Learning from In-Context Demonstrations} \\
\midrule
DeepSeek-V3 & 0.63 & 0.64 & 0.63 & 0.57 & 0.73 & 10.60 & 3.19 \\
GPT-4o & 0.77 & 0.74 & 0.70 & 0.66 & 0.76 & 7.00 & 3.63 \\
Qwen-Max & 0.48 & 0.00 & 0.00 & 0.49 & 0.00 & 13.40 & 0.96 \\
DemoDiff (Ours) & \textbf{0.81} & \textbf{0.82} & \textbf{0.78} & 0.69 & \textbf{0.79} & \textbf{2.00} & \textbf{3.90} \\
\bottomrule
\end{tabular}
\end{table}

\subsection{Details on Case Studies}\label{addsec:case-study}

\begin{figure*}[t]
    \centering
    \includegraphics[width=0.8\textwidth]{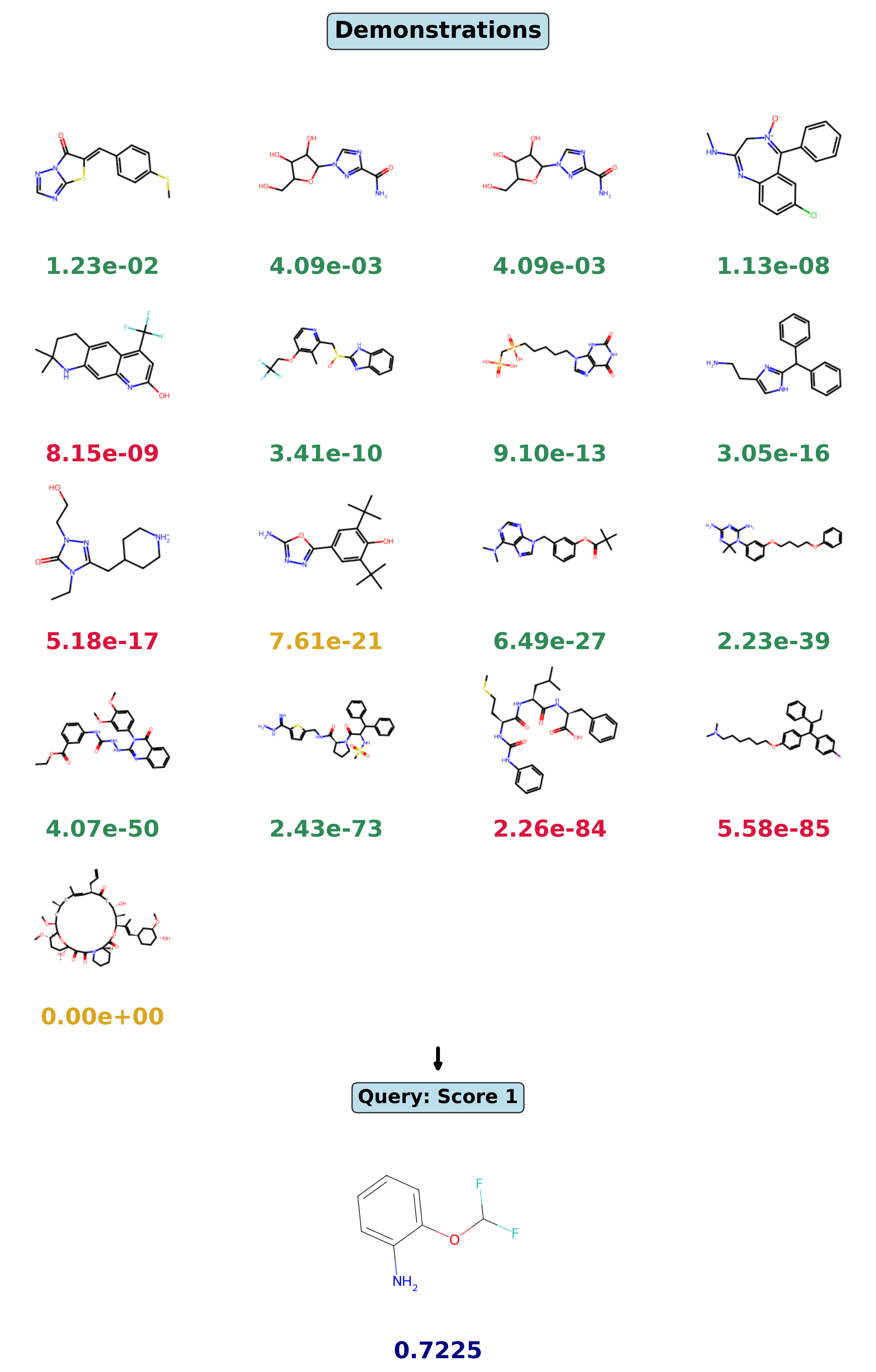}
    \caption{Structure constrained generation for Isomer with all negative demonstrations.}
    \label{fig:sub-structure}
\end{figure*}

\begin{figure*}[t]
    \centering
    \includegraphics[width=0.8\textwidth]{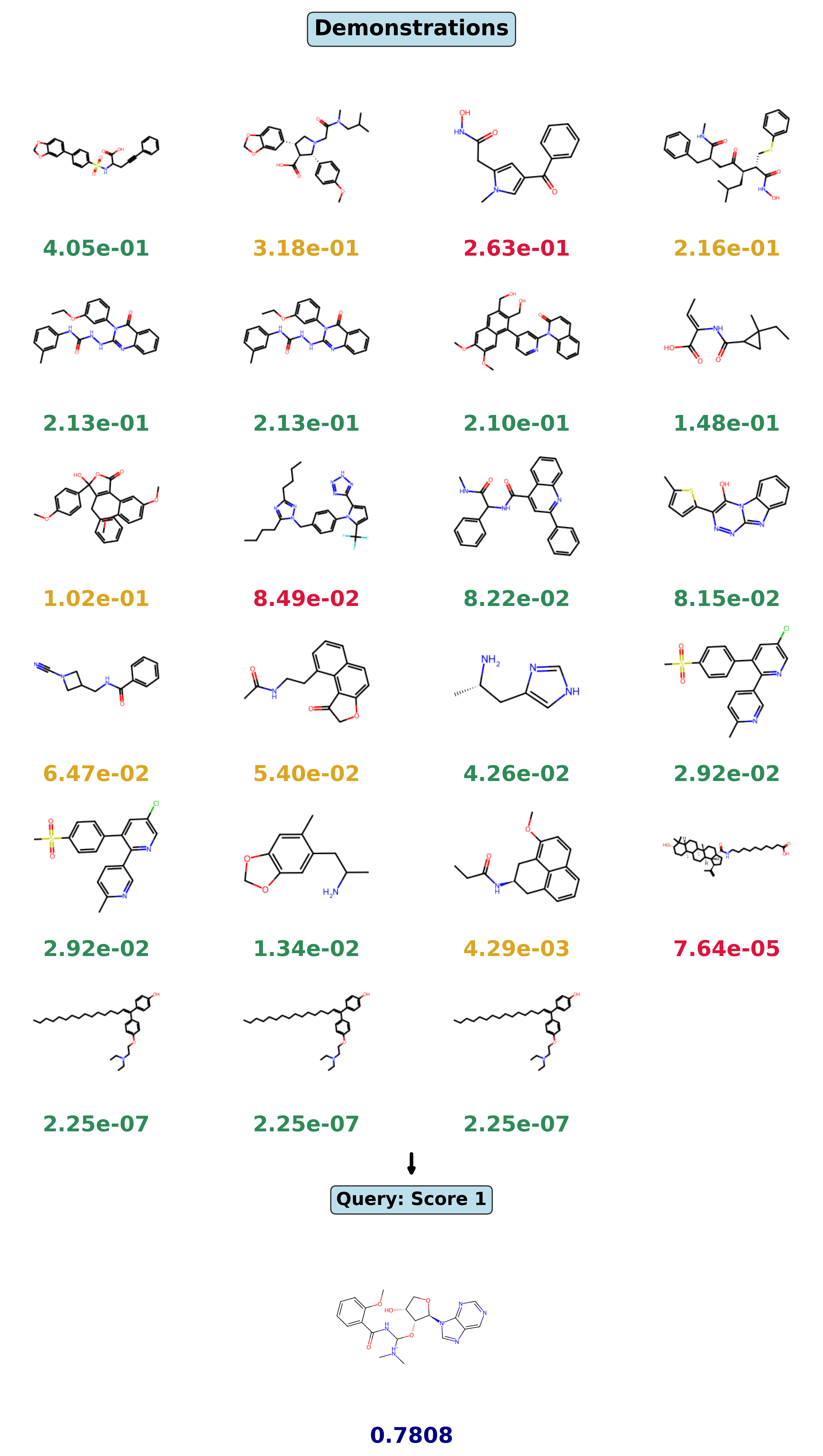}
    \caption{Drug MPO for Osimertinib with all negative demonstrations.}
    \label{fig:sub-mpo}
\end{figure*}

\begin{figure*}[t]
    \centering
    \includegraphics[width=0.8\textwidth]{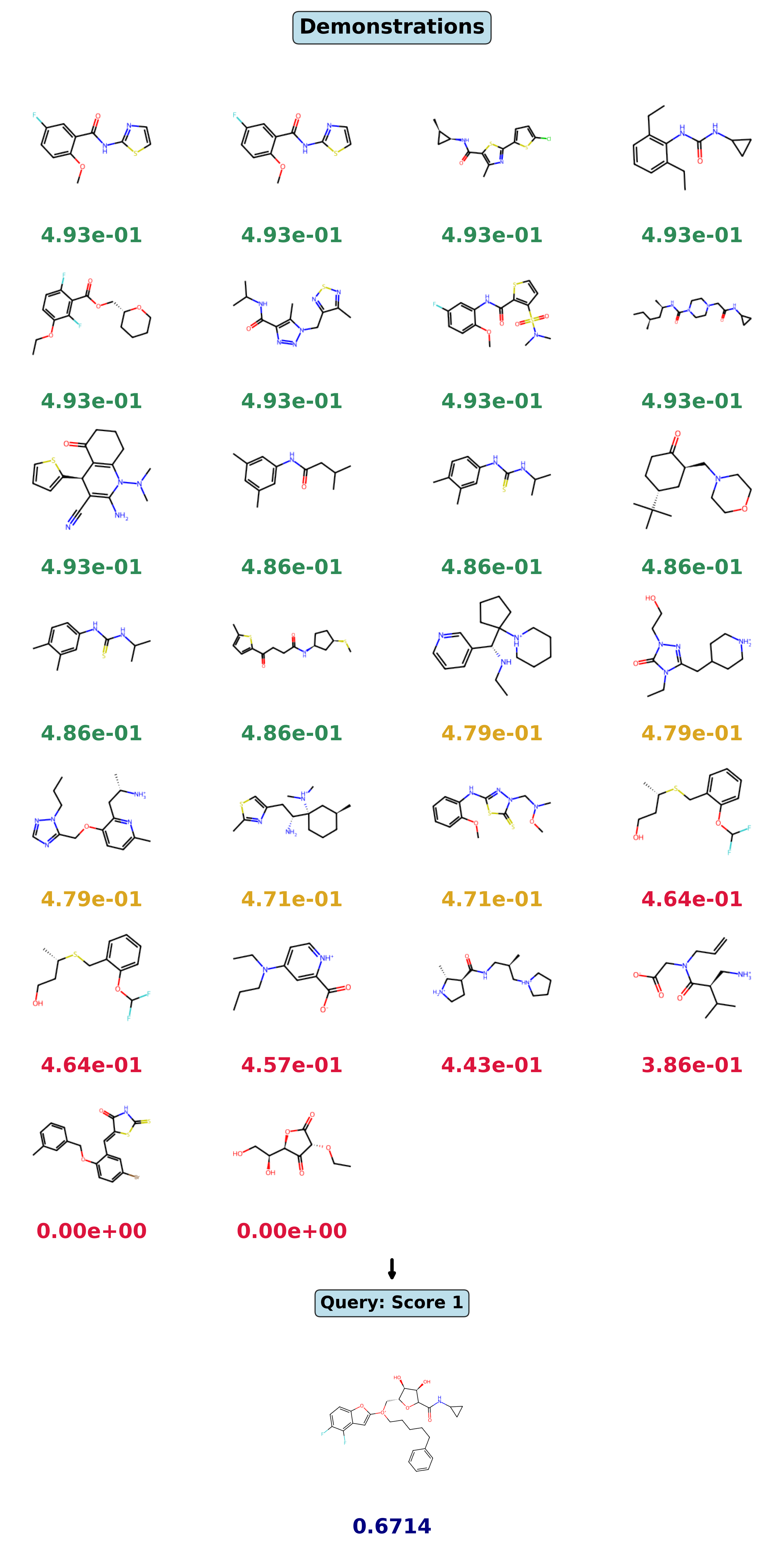}
    \caption{Target-based design for PARP1 with all negative demonstrations.}
    \label{fig:sub-target}
\end{figure*}

\cref{fig:sub-structure,fig:sub-mpo,fig:sub-target} present case studies using negative demonstrations to generate molecules with positive scores. The tasks include structure-constrained design of an isomer with 17 demonstrations, drug MPO of Osimertinib with 23 demonstrations, and protein target design of PARP1 with 26 demonstrations. These molecules contain 460, 631, and 481 atoms, respectively, far exceeding the 150-token context window under atom-level representations. The motif-level representation efficiently encodes all demonstrations within the same window.

\end{document}